%% file: WACV_2027.tex
\documentclass[10pt,twocolumn,letterpaper]{article}

\usepackage{tabularx} 
\usepackage{subcaption}
\usepackage[pagenumbers]{wacv} 
\usepackage{subcaption}
\usepackage[most]{tcolorbox}
\usepackage{graphicx}
\usepackage{enumitem}

\definecolor{wacvblue}{rgb}{0.21,0.49,0.74}
\usepackage[pagebackref,breaklinks,colorlinks,allcolors=wacvblue]{hyperref}

\def\confName{WACV}
\def\confYear{2027}

\title{Cross-Contextual Vision-Language Adaptation with LoRA for Personalized Severe Adverse Event Detection in Clinical Wound Monitoring}

\author{
Aditi Naiknaware$^1$, Jian Sun$^2$, Aminreza Khandan$^2$,\\
Shengyang Huang$^2$, Sean Dow$^3$, Bijan Najafi$^2$, and Salimeh Sekeh$^1$\\[4pt]
$^1$San Diego State University\\
$^2$University of California, Los Angeles\\
$^3$University of California, Davis\\[4pt]
{\tt\scriptsize \{anaiknaware7153,ssekeh\}@sdsu.edu}\\
{\tt\scriptsize \{JianSun,AKhandan,bnajafi\}@mednet.ucla.edu}\\
{\tt\scriptsize ovinhuang233@g.ucla.edu, sddow@ucdavis.edu}
}
\begin{document}
\maketitle
\input{sec/0_abstract}

\input{sec/1_intro}

\input{sec/2_related_work}
\input{sec/3_methodology}

\input{sec/4_experiments}

\input{sec/5_conclusion}
\input{sec/Acknowledgement}
{
    \small
    \bibliographystyle{ieeenat_fullname}
    \bibliography{main}
}
\newpage

\input{sec/x_SM}

\end{document}

%% file: sec/0_abstract.tex
\begin{abstract}
Wound monitoring is a critical yet underserved clinical challenge, where timely identification of severe adverse events (SAEs) such as infection, tissue deterioration, and delayed healing can significantly impact patient outcomes. While recent vision-language models (VLMs) have demonstrated strong multimodal reasoning capabilities, they often lack the domain-specific grounding required to effectively integrate wound imagery with heterogeneous clinical information and provide limited mechanisms for detecting clinically significant cases that differ from the training distribution. In this work, we present a multimodal framework for automated wound monitoring and SAE detection. Our approach leverages paired clinical notes and detailed wound descriptions capturing visual characteristics such as wound appearance, surrounding skin condition, color changes, and signs of inflammation or healing progression, which are encoded through a dual-stream Low-Rank Adaptation (LoRA) framework built upon a frozen BiomedCLIP backbone. We introduce a cross-contextual LoRA fusion mechanism that enables information exchange between clinical semantics and visual wound descriptors, producing context-aware multimodal representations without requiring full model fine-tuning. To identify personalized SAEs, we propose a wound-specific out-of-distribution (OOD) detection framework that combines semantic matching, visual typicality, caption-text alignment, and caption-visual alignment signals into a unified SAE (OOD) detection score. To further capture healing dynamics, we incorporate covariate consistency and temporal drift penalty that leverage changes in wound characteristics across patient visits. Preliminary experiments on a longitudinal wound dataset collected through clinical visits demonstrate promising performance on both wound healing assessment and SAE detection. These findings highlight the potential of semantically enriched, temporally aware vision-language systems for clinical wound monitoring and early risk identification.

\end{abstract}

%% file: sec/1_intro.tex
\section{Introduction}
\label{sec:intro}

\begin{figure}[t!]
\centering

\begin{subfigure}{0.48\columnwidth}
    \centering
    \includegraphics[width=0.8\linewidth]{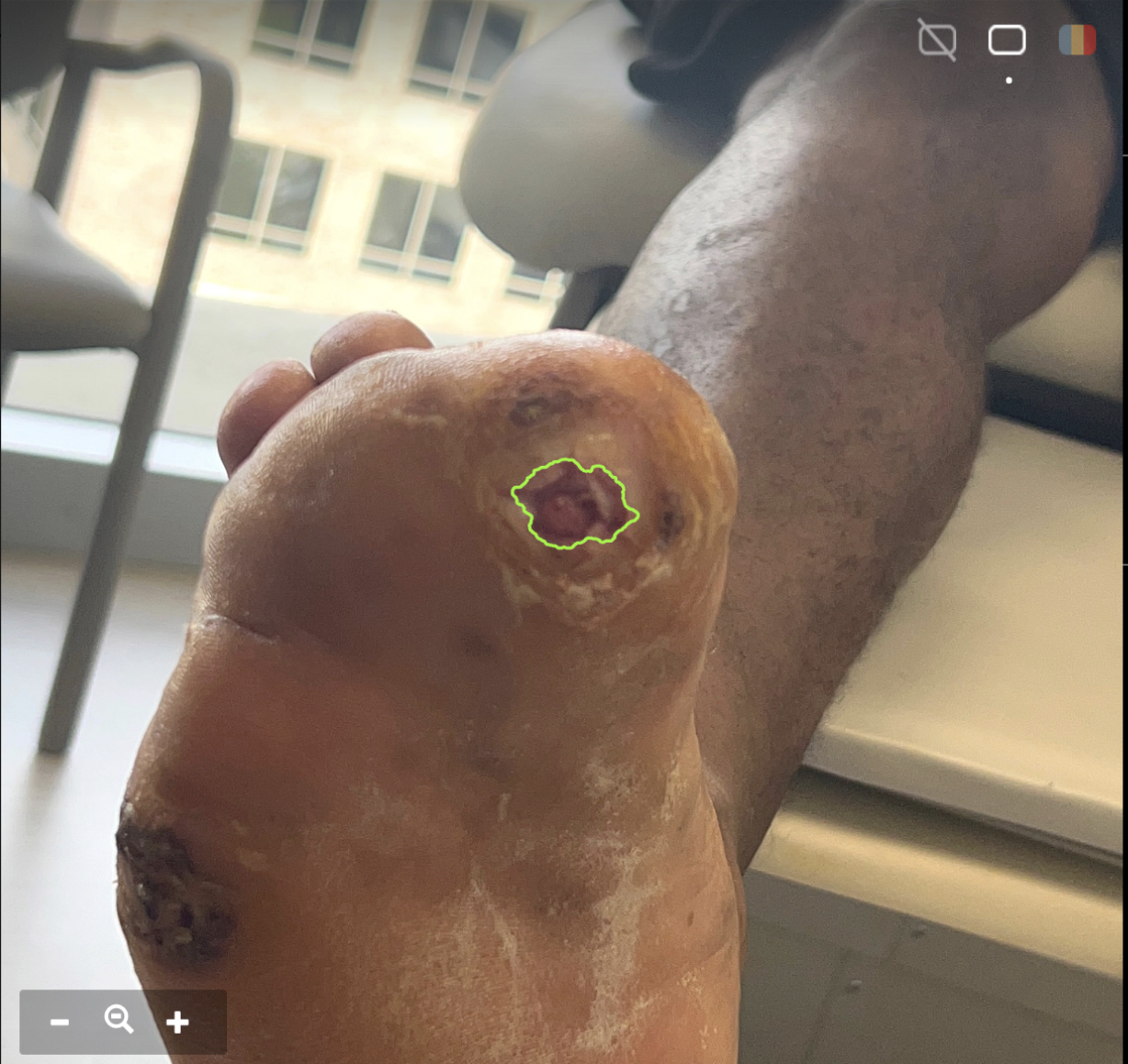}
    \caption{Healing wound}
\end{subfigure}
\hfill
\begin{subfigure}{0.48\columnwidth}
    \centering
    \includegraphics[width=0.8\linewidth]{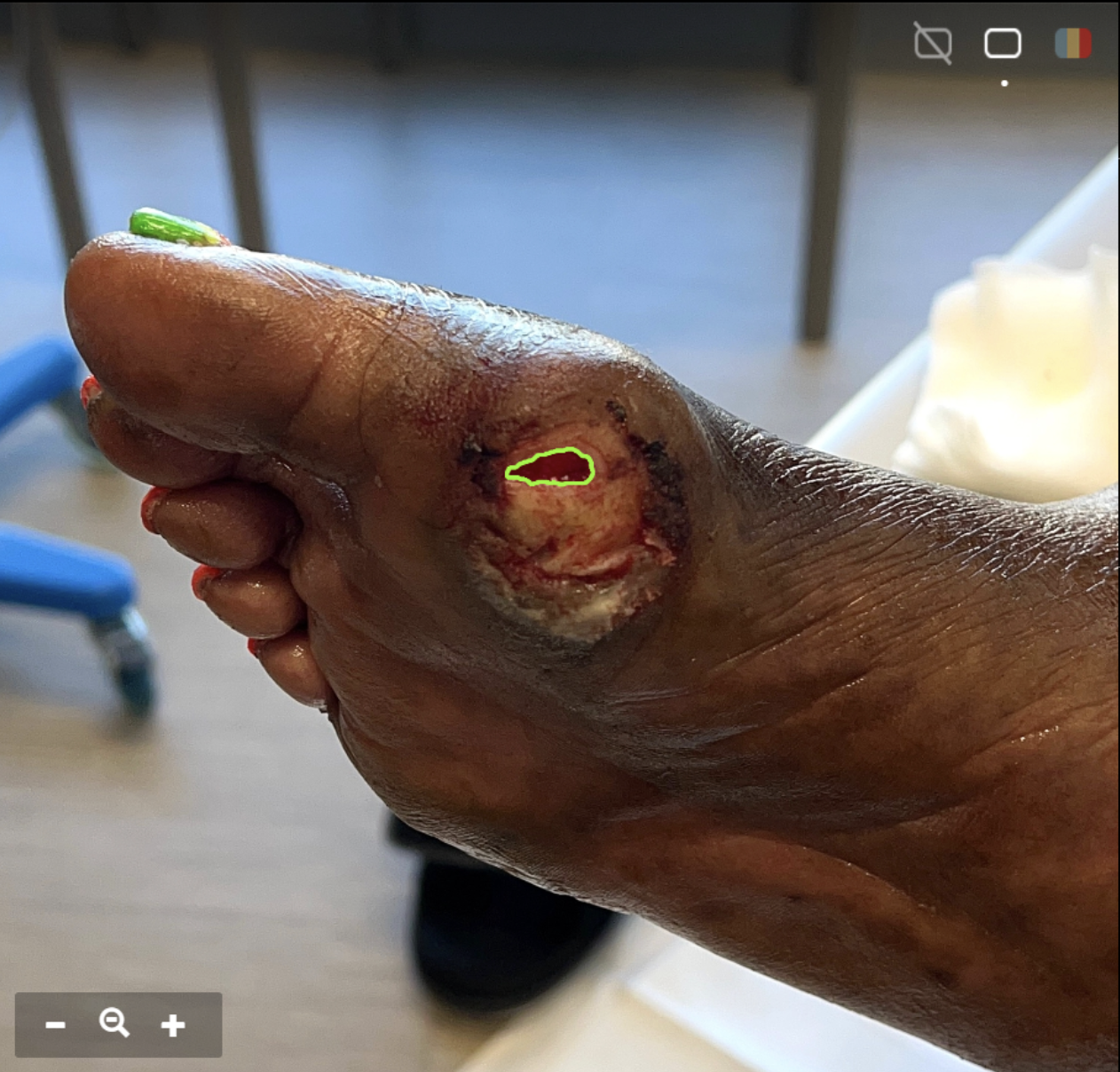}
    \caption{AE/SAE wound}
\end{subfigure}

\vspace{2mm}

\footnotesize

\begin{tabularx}{\columnwidth}{X|X}

\textbf{Healing wound} &
\textbf{AE/SAE wound} \\

\hline

\textbf{$T_c:$}

Age: 61

HbA1c: 7.2

UT Grade: 1A

&
\textbf{$T_c:$}

Age: 63

HbA1c: 9.6

UT Grade: 2B

\\

\hline

\textbf{$T_d$}: \footnotesize{The wound has healthy red tissue and appears to be healing well. The surrounding skin looks intact with only mild redness and very little dead tissue.}
&
\textbf{$T_d$} : \footnotesize{The wound contains a mix of healthy and yellow tissue with redness and swelling around it. The wound does not appear to be healing normally and shows signs of delayed healing.}

\\

\end{tabularx}

\caption{
Illustrating the need for multimodal reasoning. Although both wounds exhibit similar visual appearance and comparable wound size, their associated structured clinical information ($T_c$) and descriptive wound characteristics ($T_d$) indicate markedly different clinical trajectories. This ambiguity motivates integrating cross-contextual information beyond visual appearance alone.
}
\label{fig:motivation}
\end{figure}

Wound complications such as infection and tissue necrosis can rapidly progress to limb loss, yet current care relies on periodic visits that may miss critical changes between assessments~\cite{boulton2018}. Automated wound monitoring therefore requires not only accurate visual representation learning but also reliable detection of clinically significant complications and atypical healing trajectories. Because these events are personalized, rare, heterogeneous, and inconsistently annotated, supervised classification alone is often insufficient. Early complication detection can instead be framed, in part, as an out-of-distribution (OOD) detection problem, identifying wound presentations that deviate from expected healing patterns. Vision-language models (VLMs) provide a promising framework by integrating wound images, clinical notes, and structured visual descriptors, enabling multimodal reasoning that remains underexplored in existing wound analysis systems~\cite{radford2021clip, zhang2023biomedclip, li2023blip2, sekeh2026crossmodal, scebba2022wound}.
























Pretrained VLMs have been adapted to specialized medical domains through full fine-tuning~\cite{zhang2023biomedclip}, prompt learning~\cite{zhou2022coop, zhou2022cocoop}, and parameter-efficient methods such as Low-Rank Adaptation (LoRA)~\cite{hu2022lora}, which offers an effective balance between adaptation performance and computational cost. However, temporal modeling of longitudinal wound progression—a key cue for distinguishing normal healing from clinically significant deterioration—remains largely unexplored in multimodal OOD frameworks~\cite{naiknaware2026tqpm}. As illustrated in Figure~\ref{fig:motivation}, wounds with similar visual appearance can follow markedly different clinical trajectories, making image-based assessment alone insufficient. Integrating structured clinical information and free-text wound descriptions is therefore essential for identifying clinically significant complications, motivating a framework that jointly reasons across all modalities.

We propose a multimodal wound monitoring framework with three key contributions. First, we introduce a dual-stream cross-contextual LoRA fusion module built on a frozen BiomedCLIP backbone~\cite{zhang2023biomedclip}, which encodes clinical notes and structured wound descriptions through separate LoRA pathways with bidirectional cross-attention to produce enriched multimodal representations without full fine-tuning. Second, we develop a wound-specific OOD detector that combines semantic matching, visual typicality, caption-text alignment, and caption-visual alignment into a unified SAE score for identifying adverse wound presentations at inference. Third, we incorporate covariate consistency and temporal drift regularization to enforce cross-visit representation consistency and model longitudinal wound progression. We evaluate the framework on the longitudinal SmartBoot DFU dataset (NCT04460573), collected with the eKare inSight 3D imaging system. Our approach outperforms single-stream baselines and consistently improves detection of rare wound presentations over unimodal OOD methods. Results show that cross-contextual LoRA fusion outperforms single-stream baselines, and our multi-signal OOD score consistently surpasses unimodal detectors on rare wound presentations.

\textit{To the best of our knowledge, we are the first to address OOD detection in clinical wound monitoring with real data. Our contributions are as follows:}
\begin{itemize}[nosep,leftmargin=*,topsep=2pt]
  \item We propose a novel personalized multimodal wound monitoring framework that jointly leverages wound images, clinical notes, and structured wound descriptions for comprehensive wound assessment.
  \item We develop a dual-stream LoRA architecture with cross-contextual fusion on a frozen BiomedCLIP backbone, enabling efficient adaptation and enhanced interaction between visual characteristics and clinical semantics.
  \item We introduce a wound-specific OOD-based SAE detection strategy that identifies clinically significant complications without requiring SAE labels during training.
  \item We incorporate longitudinal wound dynamics through covariate consistency and temporal drift regularization, improving robustness for real-world wound monitoring across multiple visits.
  \end{itemize}

%% file: sec/2_related_work.tex
\section{Related Works}

Diabetic foot ulcers (DFUs) carry a high risk of amputation when healing stalls~\cite{armstrong2017diabetic, boulton2018}, yet existing automated wound-monitoring systems remain largely discriminative and cannot detect distributional anomalies associated with severe adverse events (SAEs). Prior work has focused on wound segmentation~\cite{wang2020woundseg, scebba2022wound}, tissue classification~\cite{rostami2021ensemble}, DFU benchmarks~\cite{cassidy2022dfuc2021}, and multimodal fusion of wound images with clinical metadata~\cite{anisuzzaman2022multimodal, cruciani2025dmwat}, but none explicitly addresses unsupervised detection of abnormal healing trajectories.

Our approach builds on BiomedCLIP~\cite{zhang2023biomedclip}, extending its multimodal representations with insights from BioViL-T~\cite{bannur2023biovilt}, MedCLIP~\cite{wang2022medclip}, and continual VLM adaptation~\cite{sekeh2026crossmodal}. We adapt BiomedCLIP using dual-stream Low-Rank Adaptation (LoRA)~\cite{hu2022lora}, with separate adapters for clinical context and wound descriptions fused through cross-adapter composition, while prompt tuning~\cite{zhou2022coop, zhou2022cocoop} serves as a parameter-efficient baseline.

Classical OOD detection relies on confidence scores~\cite{hendrycks2017msp, liang2018odin, liu2020energy}, auxiliary supervision~\cite{hendrycks2019deep}, activation shaping~\cite{sun2021react, djurisic2023ash}, or feature-space methods~\cite{lee2018mahalanobis, sun2022knnood, zhang2022vim}. More recent vision-language approaches exploit image-text alignment for multimodal OOD detection~\cite{ming2022delving, zhang2024vision, miyai2025glmcm, wang2023clipn, miyai2023locoop, jiang2024neglabel}. Our framework extends T-QPM~\cite{naiknaware2026tqpm} to longitudinal wound monitoring by replacing handcrafted prompts with LoRA-derived prototypes and introducing wound-area-aware temporal regularization. Additional related work is provided in the Supplementary Materials.

%% file: sec/3_methodology.tex
\section{Methods}

{\bf Problem Statement:}
The data model $\mathcal{D}=\{{\bf x}_t,y_i\}_{i=1}^n$ from wound environment $\mathcal{E}$ is given. In environment $\mathcal{E}$, each data consist of triple stream input modality: RGB image (vision) and dual-text (language modality) samples including clinical context and visual grounded wound description, $\mathcal{D}=\{\mathcal{I},\mathcal{T}\}$, where $\mathcal{I}=\{I_{i},y_{i}\}_{i=1}^{n}$ and $\mathcal{T}=\{(T^c_{i}, T^d_i)\}_{i=1}^{n}$. This means that each image $I_i$ is accompanied with a pair of texts: clinical text $T^c_{i}$ and image description $T^d_{i}$. At each time step $t$, our model receives a fused of in-distribution (ID) $\mathcal{D}^{id}$, covariate shifted $\mathcal{D}^{cov}$, and SAE (OOD) $\mathcal{D}^{ood}$ inputs: 
$\mathcal{D}^{id}=\{\mathcal{I}^{id},\mathcal{T}^{id}\}$, $\mathcal{D}^{cov}=\{\mathcal{I}^{cov},\mathcal{T}^{id}\}$, and $\mathcal{D}^{ood}=\{\mathcal{I}^{ood},\mathcal{T}^{ood}\}$. Note that $\mathcal{T}^{id}=(\mathcal{T}^{id,c},\mathcal{T}^{id,d})$ and $\mathcal{T}^{ood}=(\mathcal{T}^{ood,c},\mathcal{T}^{ood,d})$, where $(\mathcal{T}^{id,c},\mathcal{T}^{id,d})$ are clinical and wound image description of ID images and $(\mathcal{T}^{ood,c},\mathcal{T}^{ood,d})$ are clinical and wound image description of OOD images. The covariate-shifted data refers to wound images that belong to the same label space as the training data but differ due to changes in the input distribution such as noise, blur, and rotated images. 
At inference step, new data ($\mathcal{D}_{test}$) is received. Our goal is to leverage the cross-contextual information of clinical context and wound image description and predict $\mathcal{D}_{test}$'s label if it is ID or flag $\mathcal{D}_{test}$ as SAE if it is OOD.  
\begin{figure*}[t!]
   \centering
    \includegraphics[width=\linewidth]{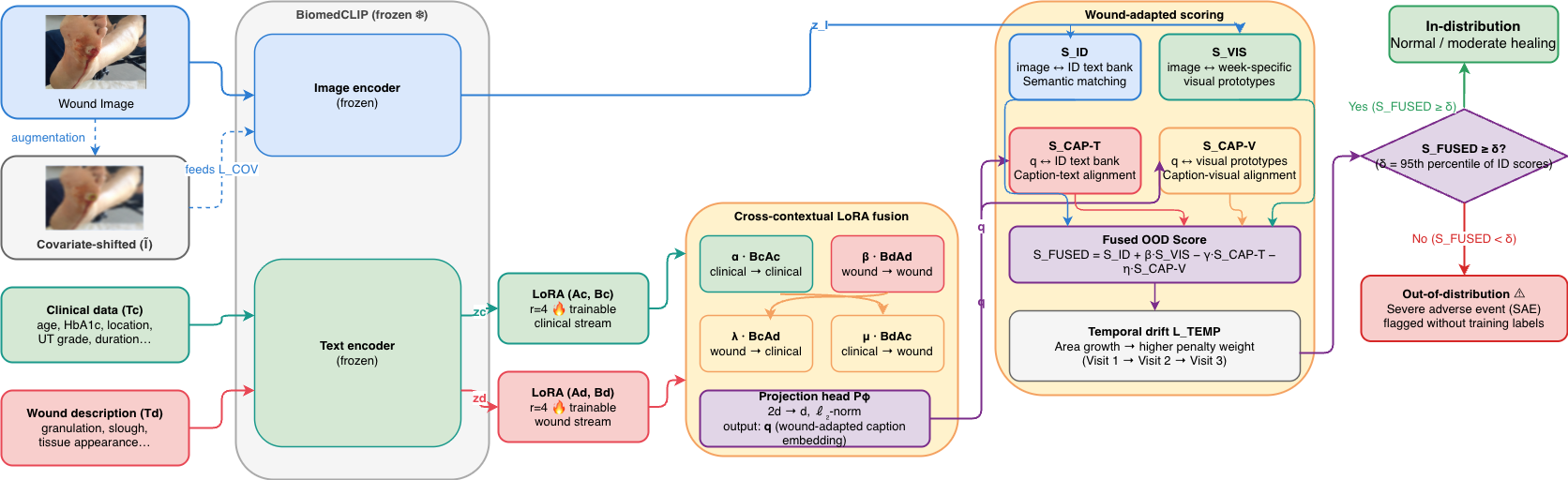}
    \caption{{ Proposed wound-adaptive multimodal OOD detection pipeline integrating frozen BiomedCLIP encoders, dual-stream LoRA adaptation, cross-contextual feature fusion, multimodal semantic scoring, and temporal drift-aware OOD classification.}}
    \label{fig:overview} 
\end{figure*}
\subsection{Personalized Low Rank Adapter Module}\label{LoRA_module}
\noindent\underline{\it Dual-Stream Text Encoding.} Each text modality is encoded independently using a shared pre-trained text
encoder $\mathcal{F}$ with frozen weights $W_0 \in \mathbb{R}^{d \times d}$. For sample $i$, we obtain two separate embedding vectors $\mathbf{x}^c_i = \mathcal{F}(T^c_i) \in \mathbb{R}^{d}$ and $\mathbf{x}^d_i = \mathcal{F}(T^d_i) \in \mathbb{R}^{d}$,  
where $\mathbf{x}^c_i$ captures the clinical context (patient
history, risk factors, medical terminology) and $\mathbf{x}^d_i$ captures
the visual-grounded wound description (tissue type, wound geometry, exudate).
Encoding the two streams separately preserves their individual semantics and
gives us explicit control over cross-stream fusion in the sequel.  

\noindent\underline{\it LoRA Adapters for Domain-Specific Text Streams.}
Standard full fine-tuning of $\mathcal{F}$ on each text stream would require updating all $d^2$ parameters of every weight matrix.
Instead, we employ Low-Rank Adaptation
(LoRA)~\cite{hu2022lora} to introduce lightweight, task-specific corrections while keeping $W_0$ frozen. For a weight matrix $W_0 \in \mathbb{R}^{d \times k}$ inside $\mathcal{F}$,
LoRA re-parameterization updates the weights as $ W = W_0 + \gamma \cdot \Delta W$, where $\Delta W = B A$, 
$A \in \mathbb{R}^{r \times k}$ and
$B \in \mathbb{R}^{d \times r}$ are trainable low-rank matrices with rank
$r \ll \min(d, k)$, and $\gamma = \alpha / r$ is a scaling factor
controlled by hyperparameter $\alpha$.
$A$ is initialized with a random Gaussian and $B$ is initialized to zero,
so $\Delta W = 0$ at the start of training, preserving the pretrained
behavior. 
We train two separate LoRA adapters on the text streams:
$$
  \text{Clinical adapter:} 
    \Delta W^c = B_c A_c, \;
     A_c \in \mathbb{R}^{r \times k},\;
     B_c \in \mathbb{R}^{d \times r},$$
$$
  \text{Wound adapter:} 
    \Delta W^d = B_d A_d, \;
     A_d \in \mathbb{R}^{r \times k},\;
     B_d \in \mathbb{R}^{d \times r}.
$$
 
\noindent Intuitively, $A_c$ learns to detect clinically relevant directions
in the input (e.g.\ severity signals, medication references), while $B_c$
maps those signals to clinical-style output corrections.
Symmetrically, $A_d$ detects descriptive grounded directions (e.g.\ necrosis,
wound geometry) and $B_d$ maps them to descriptive corrections. 

The adapted encoder outputs for sample $i$ are: 
$$
  \mathbf{h}^c_i
    = W_0\,\mathbf{x}^c_i + \gamma\, B_c A_c\,\mathbf{x}^c_i, \;\;
  \mathbf{h}^d_i
    = W_0\,\mathbf{x}^d_i + \gamma\, B_d A_d\,\mathbf{x}^d_i.
$$

 
\noindent\underline{\it Cross-Contextual LoRA Fusion.} The representations $\mathbf{h}^c_i$ and $\mathbf{h}^d_i$ capture each
modality independently, but do not yet encode the \emph{cross-contextual}
relationship between clinical context and wound description.
We therefore propose a cross-adapter fusion that explicitly couples the two streams through their LoRA components. The matrix $A_c$ has been trained to project inputs into a subspace that is
maximally informative for clinical reasoning. The matrix $B_d$ has been trained to produce wound-description-style output
corrections. Composing them as $B_d A_c$ yields an adapter that \emph{perceives the input
through clinical eyes but responds in descriptive language} and vice versa for $B_c A_d$.
This cross-composition is well-defined because both $A_\cdot \in
\mathbb{R}^{r \times k}$ and $B_\cdot \in \mathbb{R}^{d \times r}$ share the same bottleneck dimension $r$.


 
\noindent\underline{\it Merged adapter and fused representation.} Given the two independently encoded vectors, we form the concatenated
representation $\mathbf{x}_i = \bigl[\mathbf{x}^c_i \;;\; \mathbf{x}^d_i\bigr]
  \in \mathbb{R}^{2d}$. 
Next, we define the full cross-contextual merged adapter as 
\begin{equation}
  \boxed{
  \Delta W_{\mathrm{merged}}=
    \alpha\, B_c A_c
    +\beta\, B_d A_d
    + \lambda\, B_c A_d
    + \mu\, B_d A_c,
  }
  \label{eq:merged_adapter}
\end{equation}
 
\noindent where $\alpha, \beta \in \mathbb{R}_{+}$ control the strength of
the same-task terms and $\lambda, \mu \in \mathbb{R}_{+}$ control the
cross-task terms.
The four components have distinct semantic roles (Table~\ref{tab:adapter_roles}).
\begin{table}[h]
  \centering
  \footnotesize
  \caption{Semantic roles of each term in 
           $\Delta W_{\mathrm{merged}}$.}
  \label{tab:adapter_roles}
  \setlength{\tabcolsep}{4pt}
  \begin{tabularx}{\columnwidth}{@{}cXXX@{}}
    \toprule
    Term & Detects (via $A$) & Corrects (via $B$) & Semantic role \\
    \midrule
    $\alpha\, B_c A_c$ & Clinical signals   & Clinical style   & Pure clinical understanding \\
    $\beta\,  B_d A_d$ & Wound descriptors  & Wound style      & Pure wound description \\
    $\lambda\, B_c A_d$ & Wound descriptors & Clinical style   & Wound features $\to$ clinical framing \\
    $\mu\,    B_d A_c$ & Clinical signals   & Wound style      & Clinical context $\to$ visual grounding \\
    \bottomrule
  \end{tabularx}
\end{table}

The merged output for sample $i$ is 
\begin{align}\label{eq:fused_expanded}
\mathbf{h}_i
  \;=\; W_0\,\mathbf{x}_i
        \;+\; \gamma\,\Delta W_{\mathrm{merged}}\,\mathbf{x}_i.
\end{align}
        
The vector $\mathbf{h}_i \in \mathbb{R}^{2d}$ is the \emph{cross-contextually enriched text representation} that will be passed alongside the image
$I_i$ to the VLM in the downstream module.  

\subsection{ Wound-Adapted OOD Detector Module}
We now describe how the cross-contextual representation $\mathbf{h}_i \in
\mathbb{R}^{2d}$ in  (\ref{eq:fused_expanded}) produced by the personalized LoRA fusion
(Section~\ref{LoRA_module}) is used as a semantically enriched
multimodal caption in a wound-adapted instantiation of the Temporal
Quadruple-Pattern Matching (T-QPM) framework~\cite{naiknaware2026tqpm}.

Our setting differs from standard T-QPM because: 
\begin{itemize}[nosep,leftmargin=*,topsep=2pt]
  \item \textbf{Personalized approach.} Through LoRA adaptation mechanism, we develop a personalized technique.
  \item \textbf{Pre-defined class prompts.} Instead of constructing the ID text bank from hand-crafted templates, we build it directly from the fused representations of ID training samples, making the bank fully data-driven.
  \item \textbf{SAE identification.} The label space is $\mathcal{Y}=\{0,1\}$ (normal and moderate healing, both in-distribution), while class~2 (SAE) is the OOD target — the detector must flag class-2 samples at inference without ever receiving their labels during training.
  \item \textbf{Temporal component.} the temporal component is derived by wound area measurements, which provide a direct physiological signal of healing trajectory: a healing wound contracts
over time, so unexpected area growth between consecutive timesteps constitutes a temporal anomaly that amplifies the temporal penalty.
  \end{itemize}

\noindent\underline{\it Projection of Fused Representation.} Because $\mathbf{h}_i \in \mathbb{R}^{2d}$ lives in the concatenated
embedding space whereas CLIP's shared vision-language space is $\mathbb{R}^{d}$, we introduce a lightweight linear projection head $P_\phi \in \mathbb{R}^{d \times 2d}$ with trainable parameters $\phi$,
followed by $\ell_2$-normalization, to map $\mathbf{h}_i$ back into the
CLIP embedding space:
\begin{align}
  \mathbf{q}_i
  \;=\;
  \mathrm{Normalize}\!\bigl(P_\phi\,\mathbf{h}_i\bigr)
  \;\in\; \mathbb{R}^{d}.
  \label{eq:proj}
\end{align}
\noindent $\mathbf{q}_i$ is the \emph{wound-adapted caption embedding} for sample
$i$, serving as a drop-in replacement for the caption query $\mathbf{q}_c$
in all four OOD detector scoring functions defined in sequel. It encodes the joint semantics of both the clinical record $T^c_i$ and the
wound image description $T^d_i$ through the cross-contextual LoRA fusion,
providing a richer and more medically grounded signal.
The projection head $P_\phi$ is optimized jointly with the two 
fusion scalars; all other parameters remain frozen.

\noindent\underline{\it Data-Driven ID Text Bank Construction.} We construct the ID
text bank directly from the fused representations of ID training samples.
For each ID class $k \in \{0,1\}$, we compute the class prototype as the
$\ell_2$-normalised mean of the projected caption embeddings of all
training samples belonging to that class:
\begin{align}
  \mathbf{t}_k
  \;&=\;
  \mathrm{Normalize}\!\!\left(
    \frac{1}{|\mathcal{S}_k|}
    \sum_{i \in \mathcal{S}_k} \mathbf{q}_i
  \right)
  \in \mathbb{R}^{d},\;\hbox{where}\noindent\\
  \mathcal{S}_k
 & = \bigl\{i : y_i = k,\;
             (I_i, T^c_i, T^d_i) \in \mathcal{D}^{\mathrm{id}}\bigr\},
  \label{eq:text_bank}
\end{align}

\noindent
where $\mathbf{q}_i$ is computed via Eq.~\eqref{eq:proj} using the
frozen LoRA-adapted encoder and projection head.
Stacking the two prototypes yields the ID text bank
$\mathbf{T}^{{id}} = [\mathbf{t}_0, \mathbf{t}_1]^\top \in
\mathbb{R}^{2 \times d}$.

\noindent
This construction is more principled than prompt engineering for two
reasons.
First, $\mathbf{t}_k$ aggregates the actual clinical and visual semantics
of class-$k$ wound samples rather than relying on generic textual
descriptions.
Second, because $\mathbf{q}_i$ is itself the output of the cross-contextual
LoRA fusion, the text bank inherits the personalised domain adaptation
from Stage~III, ensuring that the ID prototypes and the test caption
embeddings live in the same adapted subspace and are therefore directly
comparable.$\mathbf{T}^{{id}}$ is computed once after the LoRA adapters and projection head are trained, and kept fixed across all post-operative timesteps.

\noindent\underline{\it Temporal Visual Prototype Construction.} Wound appearance evolves systematically over time, inducing a temporal
distribution shift in the visual domain.
To handle this, we compute timestep-specific visual prototypes from the
frozen CLIP-ViT encoder $\phi^V$.
For each wound image $I$ at post-operative timestep $t$, the encoder
produces patch embeddings
$\mathbf{F}(I) = \phi^V(I) \in \mathbb{R}^{(N+1)\times d}$,
decomposed into a global CLS token
$\mathbf{F}_v(I) \in \mathbb{R}^d$
and spatial patch features
$\mathbf{F}_s(I) \in \mathbb{R}^{N \times d}$.

For each ID class $k \in \{0,1\}$, class-specific spatial attention
weights are computed as:
\begin{equation}
  \mathbf{A}_k(I)
  = \mathrm{Softmax}\!\!\left(
      \frac{\mathbf{F}_s(I)\,\mathbf{t}_k}
           {\|\mathbf{F}_s(I)\|\,\|\mathbf{t}_k\|}
    \right) \in \mathbb{R}^N,
  \label{eq:spatial_attn}
\end{equation}
\noindent yielding a class-attended spatial feature
$\tilde{\mathbf{f}}_k(I) = \mathbf{A}_k(I)^\top \mathbf{F}_s(I)$.
The class-specific image representation combines global and attended
spatial features as
$\mathbf{f}_k(I) = \gamma_s\,\tilde{\mathbf{f}}_k(I) + \mathbf{F}_v(I)$.
The ID logit vector and class probability vector follow as:
\begin{align}
  \mathbf{z}_{id}(I)
  &= \bigl[\mathbf{f}_0(I)^\top\mathbf{t}_0,\;
           \mathbf{f}_1(I)^\top\mathbf{t}_1\bigr]^\top
  \in \mathbb{R}^2,\nonumber\\
  \qquad
  \mathbf{p}(I) &= \mathrm{Softmax}\!\left(\frac{\mathbf{z}_{id}(I)}{\tau}\right),
  \label{eq:id_logits}
\end{align}

\noindent
and the timestep-specific class-conditional visual prototype for class $k$
at day $t$ is $ \boldsymbol{\mu}_{k,t}
  = \mathbb{E}\bigl[\mathbf{p}(I)
    \mid (I,y) \in \mathcal{D}_t^{\mathrm{train}},\; y = k\bigr]
  \in \mathbb{R}^2$.
The set $\{\boldsymbol{\mu}_{k,t}\}_{k=0}^{1}$ defines the expected ID
probability pattern at day~$t$, allowing the notion of
``normal'' wound healing behaviour to adapt to the gradual visual changes
inherent in post-surgical recovery.

\noindent\underline{\it Wound-Adapted Quadruple Cross-Modal Scoring.} For a test wound image $I_i$ with its wound-adapted caption embedding
$\mathbf{q}_i$ (Eq.~\eqref{eq:proj}) observed at timestep
$t$, we construct four complementary OOD scores: $S_{\mathrm{ID}}$, a semantic matching score (image $\leftrightarrow$ ID text bank), $S_{\mathrm{VIS}}$ a wound-based score (image $\leftrightarrow$ ID visual prototypes), $S_{\mathrm{CAP\text{-}T}}$ a caption-text alignment score (fused caption $\leftrightarrow$ ID text bank), and $S_{\mathrm{CAP\text{-}V}}$ a caption-visual alignment score (fused caption $\leftrightarrow$ ID visual prototypes). Details on scores computation are provided in the Supplementary Materials.

\noindent\textbf{Fused OOD Score.}
The four scores are combined with learnable positive weights
$\beta, \eta > 0$ as:
\begin{align}
  S_{\mathrm{FUSED}}(I_i,t)
  = S_{\mathrm{ID}}(I_i)
  + \beta\cdot S_{\mathrm{VIS}}(I_i,t)\nonumber\\
  - \gamma_{\mathrm{cap}}\cdot S_{\mathrm{CAP\text{-}T}}(I_i)
  - \eta\cdot S_{\mathrm{CAP\text{-}V}}(I_i,t),
\end{align}
\noindent where $\gamma_{\mathrm{cap}} > 0$ is a fixed hyperparameter and
$\beta = \log(1+e^{\tilde{\beta}})$, $\eta = \log(1+e^{\tilde{\eta}})$
are softplus-reparameterised to enforce strict positivity.
The caption-based terms are subtracted because high caption-to-ID-bank
alignment, when not matched by visual typicality, is the hallmark of
a SAE: the fused clinical-visual description superficially resembles
known healing language, yet the wound image itself departs from the
expected healing trajectory.

\subsection{Area-reweighted Total Loss Function}\label{sec:phase4}
At the initial timestep $t=0$, a global decision threshold
$\delta$ is calibrated as the $\delta_q$-th percentile of fused scores on
the ID training data with $y\in\{0,1\}$:
\begin{equation*}
  \delta
  = \mathrm{quantile}_{\delta_q}\!\bigl(
      \{S_{\mathrm{FUSED}}(I,0)
        \mid (I,y) \in \mathcal{D}_0^{\mathrm{train}}\}
    \bigr),
  \label{eq:threshold}
\end{equation*}

\noindent
and kept fixed across all timesteps (clinical visits).

\noindent\textbf{Balanced ID classification loss.}
Cross-entropy is computed symmetrically on both clean and
covariate-shifted wound views over the two ID classes:
\begin{align}
  \mathcal{L}_{\mathrm{ID}}
  = \frac{1}{2}\Bigl(
      \mathbb{E}_{(I,y)\sim\mathcal{D}_t^{\mathrm{id}}}
        \bigl[\mathcal{L}_{\mathrm{CE}}(\mathbf{z}_{id}(I),\,y)\bigr]\nonumber\\
      +
      \mathbb{E}_{(\tilde{I},y)\sim\mathcal{D}_t^{\mathrm{id}}}
        \bigl[\mathcal{L}_{\mathrm{CE}}(\mathbf{z}_{id}(\tilde{I}),\,y)\bigr]
    \Bigr),
  \label{eq:l_id}
\end{align}
\noindent
where $\tilde{I}$ is a covariate-shifted view of $I$ (e.g., lighting
perturbation, image compression), and the expectation is taken only over
ID samples $y \in \{0,1\}$.

\noindent\textbf{Covariate consistency loss.}
Enforces that OOD detection scores remain stable under covariate shift:
\begin{equation}
  \mathcal{L}_{\mathrm{COV}}
  = \mathbb{E}_{I\sim\mathcal{D}_t^{\mathrm{id}}}\Bigl[
      \bigl|S_{\mathrm{FUSED}}(I,t)
            - S_{\mathrm{FUSED}}(\tilde{I},t)\bigr|
    \Bigr].
  \label{eq:l_cov}
\end{equation}
\noindent\textbf{Area-reweighted temporal drift penalty.}
Wound area should decrease monotonically under normal recovery.
We exploit this domain knowledge to define a temporally-aware penalty that
is reweighted by the observed wound area change between consecutive days. For sample $i$ at timestep $t$, let $a_i^t \in \mathbb{R}_{+}$ denote
the measured wound area (e.g., in cm$^2$).
The signed area change is:
\begin{equation}
  \Delta a_i^t = a_i^t - a_i^{t-1},
  \label{eq:delta_area}
\end{equation}
\noindent
where $\Delta a_i^t < 0$ indicates contraction (expected healing) and
$\Delta a_i^t > 0$ indicates expansion (physiological anomaly).
We normalize the area change across the training batch at each timestep
to obtain a unit-scale reweighting signal:
\begin{equation}
  w_i^t
  = \sigma\!\left(
      \frac{\Delta a_i^t - \mu_{\Delta}}{\sigma_{\Delta}}
    \right),
  \label{eq:area_weight}
\end{equation}
\noindent
where $\mu_{\Delta}$ and $\sigma_{\Delta}$ are the batch mean and standard
deviation of $\{\Delta a_i^t\}$, and $\sigma(\cdot)$ is the sigmoid
function.
$w_i^t \in (0,1)$ is close to~1 when the wound is expanding (anomalous)
and close to~0 when it is contracting (expected).
Samples with unexpected area growth thus receive a higher temporal penalty weight. The area-reweighted temporal drift penalty penalizes inconsistency in
the fused OOD score between consecutive timesteps, scaled by the
physiological anomaly weight:
\begin{align}
 & \mathcal{L}_{\mathrm{TEMP}}
  = \mathbb{E}_{I\sim\mathcal{D}_t^{\mathrm{id}}}\!\Bigl[
      w_i^t \cdot
      \bigl|S_{\mathrm{FUSED}}(I,t)
            - S_{\mathrm{FUSED}}(I,t-1)\bigr|
    \Bigr]\nonumber\\
&  +
  \mathbb{E}_{\tilde{I}\sim\mathcal{D}_t^{\mathrm{id}}}\!\Bigl[
      w_i^t \cdot
      \bigl|S_{\mathrm{FUSED}}(\tilde{I},t)
            - S_{\mathrm{FUSED}}(\tilde{I},t-1)\bigr|
    \Bigr],
  \label{eq:l_temp}
\end{align}
\noindent

This formulation encodes the prior that a wound whose area is growing
should produce a more dramatically shifted OOD score — and thus be
penalised more heavily if its score remains unchanged — while a wound
whose area is shrinking normally incurs little temporal penalty.

\noindent\textbf{United loss functions.}
The combined objective with Lagrangian multipliers
$\lambda_{\mathrm{cov}}, \lambda_{\mathrm{temp}} > 0$ is:
\begin{equation}
  \boxed{
  \mathcal{L}_{\mathrm{TOTAL}}
  = \mathcal{L}_{\mathrm{ID}}
  + \lambda_{\mathrm{cov}}\,\mathcal{L}_{\mathrm{COV}}
  + \lambda_{\mathrm{temp}}\,\mathcal{L}_{\mathrm{TEMP}}.
  }
  \label{eq:l_total}
\end{equation}
\noindent
At each timestep $t$, only three parameter sets are
updated: the projection head $P_\phi$ and the two fusion scalars
$\tilde{\beta}, \tilde{\eta}$.
All CLIP backbone weights and LoRA adapter weights remain frozen,
ensuring that the pretrained vision-language representations are
preserved throughout.

\noindent\textbf{OOD decision.}
Given a test wound image $I_i$ with fused caption embedding $\mathbf{q}_i$
at post-operative visits $t$:
\begin{equation}
  \mathcal{D}(I_i,t)
  =
  \begin{cases}
    \texttt{ID, class }k^*
      &\hspace{-0.5cm} \text{if}\; S_{\mathrm{FUSED}}(I_i,t) \leq \delta, \\
    \texttt{OOD (SAE flagged)}
      & \text{otherwise,}
  \end{cases}
  \label{eq:decision}
\end{equation}
\noindent
where $k^* = \arg\max_{k \in \{0,1\}}\,\mathbf{z}_{\mathrm{ID}}(I_i)[k]$
is the predicted healing class for ID samples.
Class-2 samples (SAE) are never seen during training; the model learns to
flag them purely from the mismatch between their fused score and the
ID-calibrated threshold $\delta$.

%% file: sec/4_experiments.tex
\section{Experiments}

\subsection{Wound Dataset and SAE Clinical Description}
The wound image dataset is from the SmartBoot DFU offloading study (NCT04460573), a 12-week randomized trial enrolling adults ($\geq$18\,yr) with active, non-infected, non-ischemic DFUs.
Digital wound images and wound-size measurements were collected at scheduled visits using the eKare inSight 3D system; structured clinical variables (ulcer area, duration, location, University of Texas classification, HbA1c) were linked at the participant and visit level.
A preprocessing pipeline converted longitudinal records into a temporal OOD benchmark.
Full details on temporal organization, outcome labeling,
multimodal representation, and patient-level partitioning are in
the supplementary material.

\subsection{Comparison with existing methods}

\begin{figure*}[t]
    \centering

    \begin{subfigure}{0.32\linewidth}
        \centering
        \includegraphics[width=\linewidth]{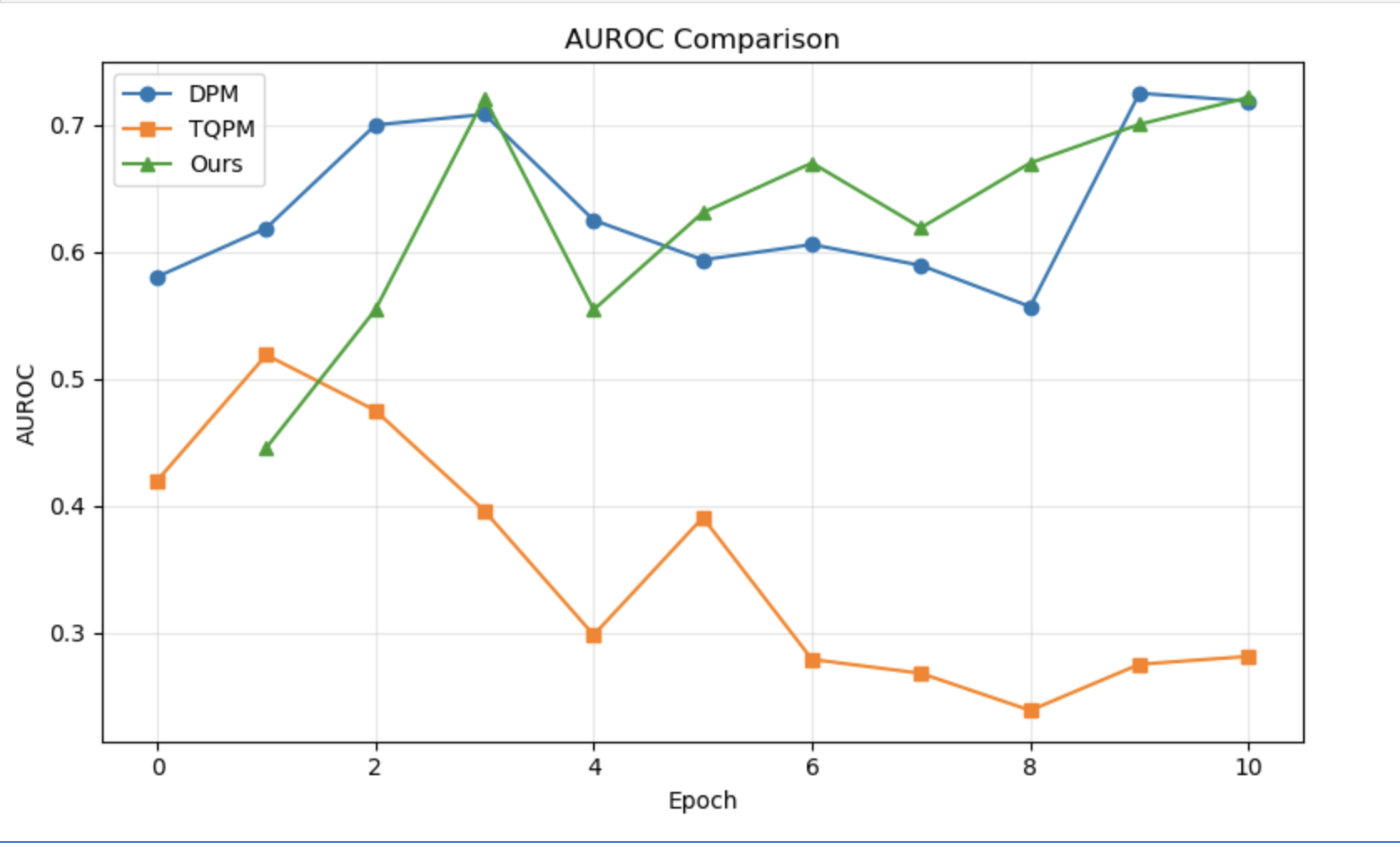}
        \caption{AUROC}
        \label{fig:auroc_comp}
    \end{subfigure}
    \hfill
    \begin{subfigure}{0.32\linewidth}
        \centering
        \includegraphics[width=\linewidth]{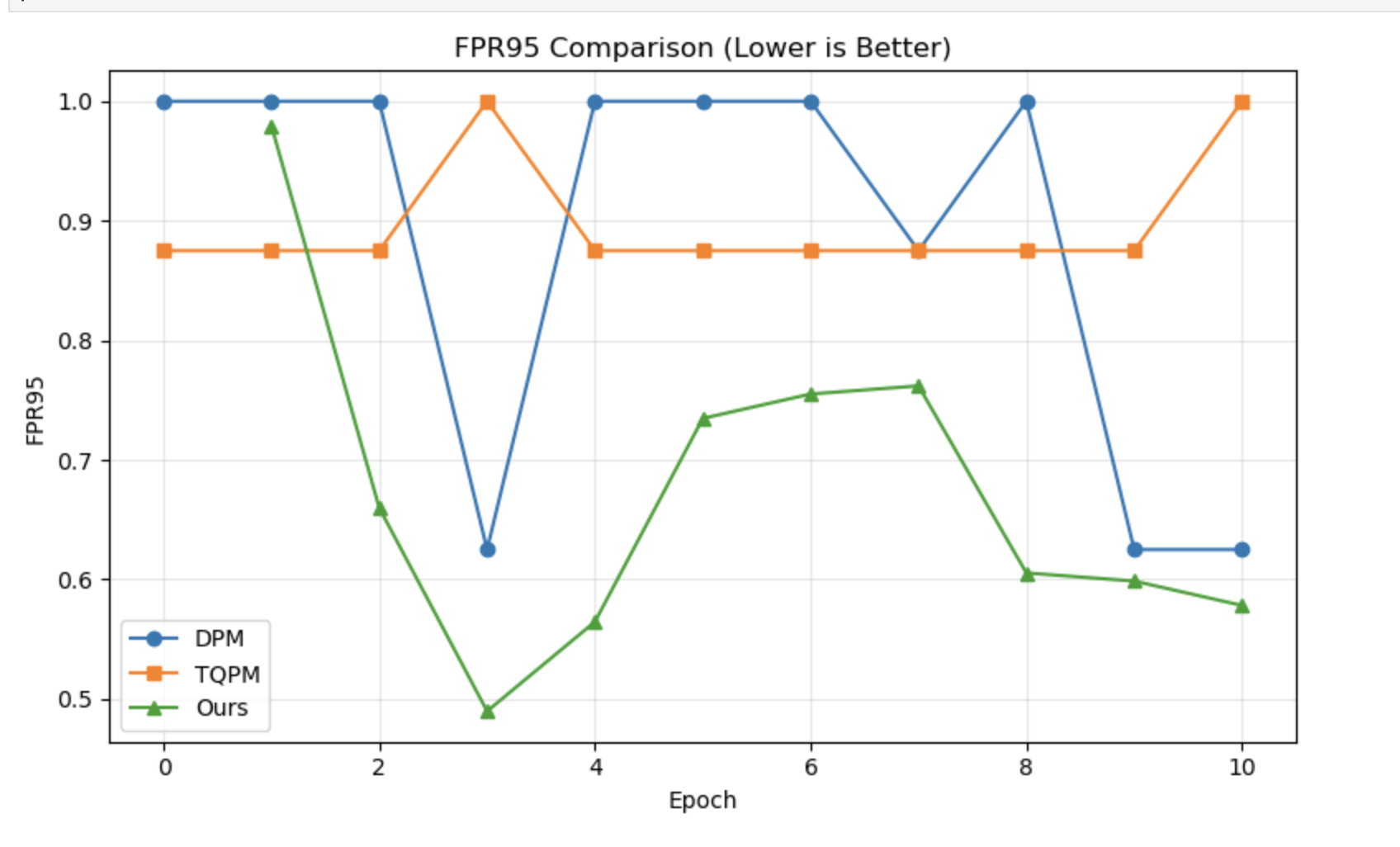}
        \caption{FPR95}
        \label{fig:fpr_comp}
    \end{subfigure}
    \hfill
    \begin{subfigure}{0.32\linewidth}
        \centering
        \includegraphics[width=\linewidth]{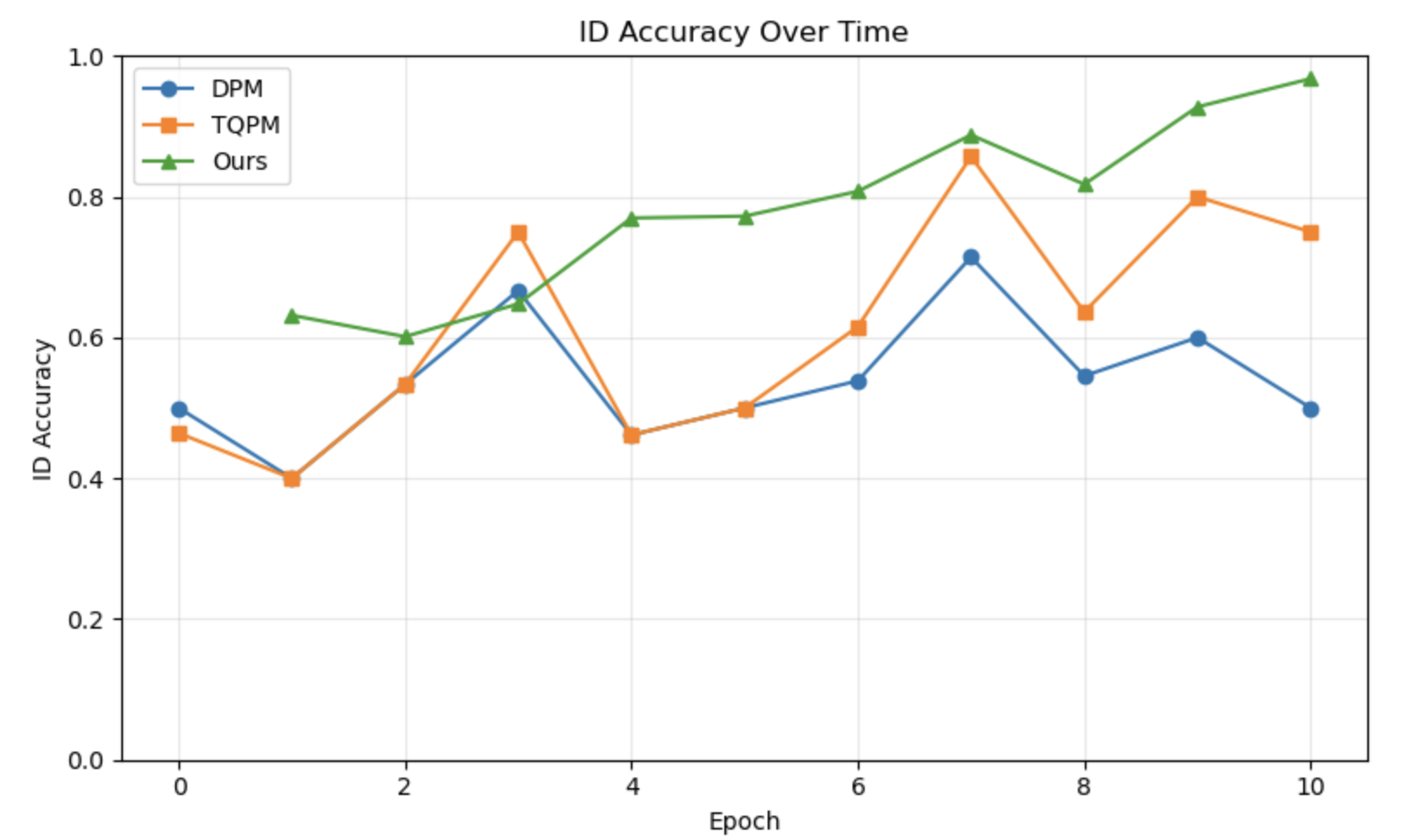}
        \caption{ID Accuracy}
        \label{fig:id_comp}
    \end{subfigure}

    \caption{
    Comparison of the proposed framework against DPM and TQPM on the SmartBoot wound benchmark.
    (a) AUROC (higher is better),
    (b) FPR95 (lower is better),
    and (c) ID Accuracy (higher is better).
    While DPM achieves a marginally higher AUROC, the proposed multimodal framework substantially reduces FPR95 and achieves the highest ID classification accuracy.
    }

    \label{fig:baseline_comparison}

\end{figure*}

We compare against DPM, TQPM, and LoCoOP baselines with the same patient-level train/val/test splits.
As shown in Table~\ref{tab:sota_comparison}, our method achieves AUROC~0.729 and the lowest FPR95 (0.490), reducing false positives by 22\% vs.\ DPM and 44\% vs.\ TQPM, while attaining the highest ID accuracy (0.937).
Training dynamics are visualized in Figure~\ref{fig:baseline_comparison}.

\begin{table}[t]
\centering
\caption{Comparison with state-of-the-art wound OOD detection methods. Best results are shown in \textbf{bold}. Higher values are better for AUROC and ID Accuracy, while lower values are preferred for FPR95.}
\label{tab:sota_comparison}

\begin{tabular}{lccc}
\hline
Method & AUROC $\uparrow$ & FPR95 $\downarrow$ & ID Accuracy $\uparrow$\\
\hline
DPM~\cite{zhang2024vision}  & 0.721 & 0.625 & 0.844 \\
LoCoOp~\cite{miyai2023locoop} & 0.662 & 0.783 & 0.839\\
TQPM~\cite{naiknaware2026tqpm} & 0.519 & 0.875 & 0.824 \\
{\bf Ours} & \textbf{0.729} & \textbf{0.490} & \textbf{0.937} \\
\hline
\end{tabular}
\end{table}

\subsection{Training Dynamics}

\begin{figure*}[t]
  \centering
  \begin{subfigure}{0.32\linewidth}
    \centering
    \includegraphics[width=\linewidth]{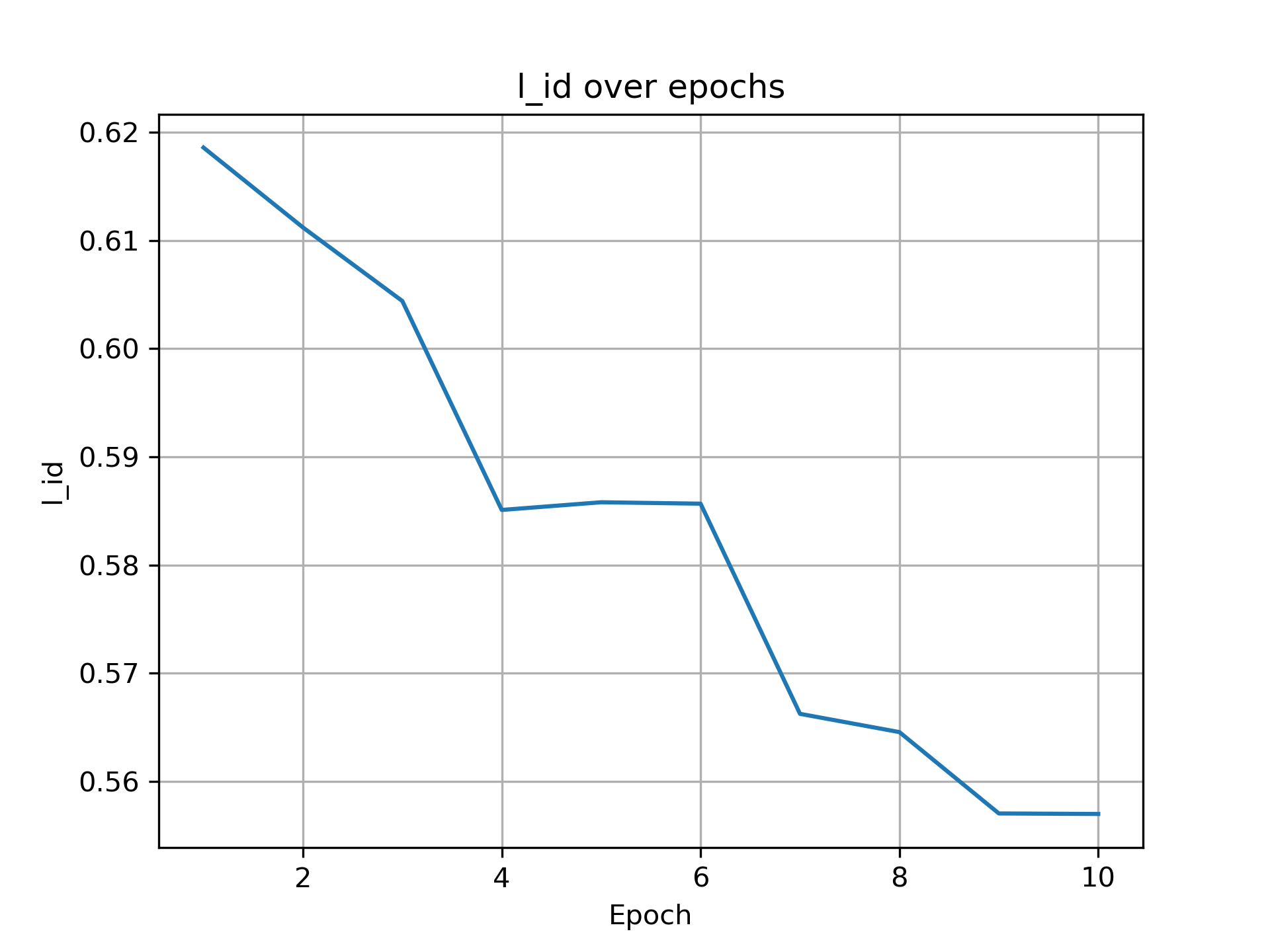}
    \caption{$L_{\mathrm{ID}}$}\label{fig:id_loss}
  \end{subfigure}\hfill
  \begin{subfigure}{0.32\linewidth}
    \centering
    \includegraphics[width=\linewidth]{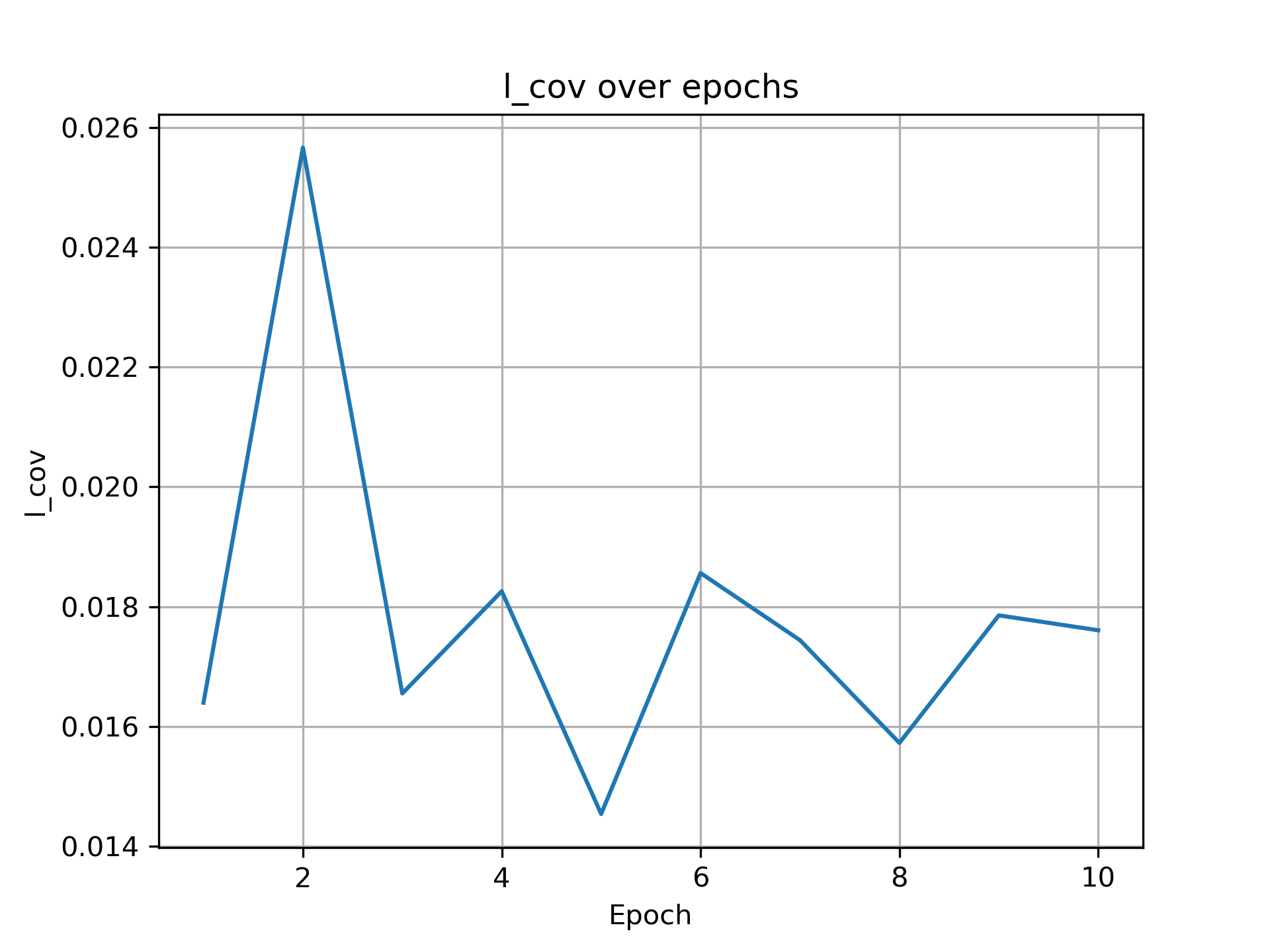}
    \caption{$L_{\mathrm{COV}}$}\label{fig:cov_loss}
  \end{subfigure}\hfill
  \begin{subfigure}{0.32\linewidth}
    \centering
    \includegraphics[width=\linewidth]{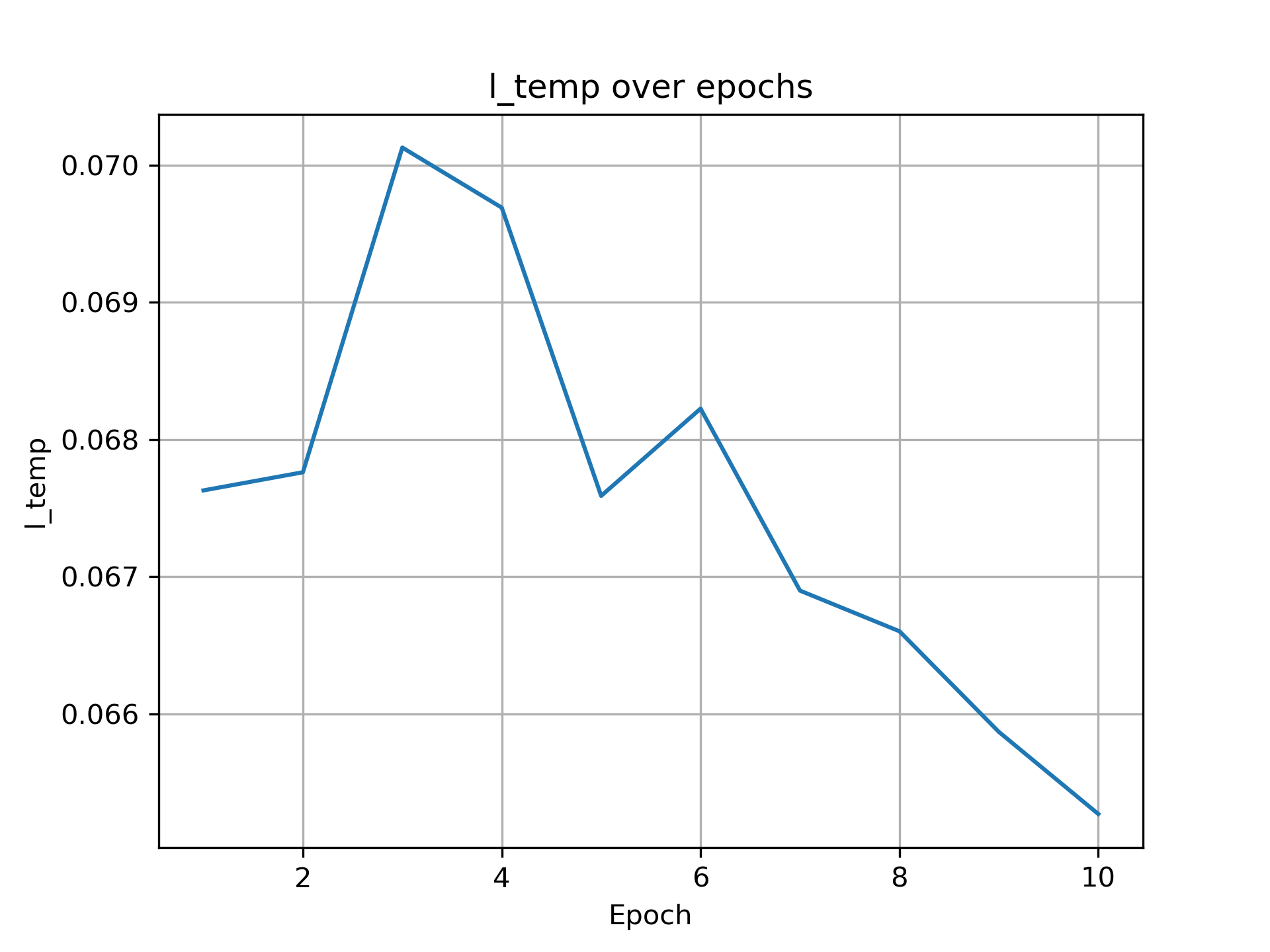}
    \caption{$L_{\mathrm{TEMP}}$}\label{fig:temp_loss}
  \end{subfigure}
  \caption{Loss curves over training.
    (a)~$L_{\mathrm{ID}}$ decreases rapidly.
    (b)~$L_{\mathrm{COV}}$ remains low under visual perturbations.
    (c)~$L_{\mathrm{TEMP}}$ decreases steadily, stabilizing longitudinal
    OOD scores.}
  \label{fig:loss_optimization_curves}
\end{figure*}

Figure~\ref{fig:loss_optimization_curves} shows the three loss components over 10 epochs.
$\mathcal{L}_{\mathrm{ID}}$ falls rapidly in early epochs, confirming stable classification adaptation. 
$\mathcal{L}_{\mathrm{COV}}$ remains low throughout, verifying covariate-shift invariance, and $\mathcal{L}_{\mathrm{TEMP}}$ decreases steadily, stabilizing the longitudinal representation.
Validation AUROC rises from 0.43 (epoch~1) to a peak of 0.75--0.76 (epochs 8--9), FPR95 drops from 0.93 to below 0.50, and ID accuracy reaches $\approx$97\% by the final epoch.

\begin{figure*}[t]
\centering
\scriptsize
\setlength{\tabcolsep}{4pt}
\renewcommand{\arraystretch}{0.75}

\newcommand{\threshold}{0.85}

\newcommand{\casecell}[6]{%
  \begin{minipage}[t]{0.32\linewidth}
    \centering
    \includegraphics[width=\linewidth,height=2.2cm,keepaspectratio]{#1}\\[1pt]
    {\footnotesize $T_c$: Age:~#3, HbA1c:~#4, UT:~#5}\\[1pt]
    {\footnotesize $T_d$: #6}\\[1pt]
    \ifdim #2pt > \threshold pt
      {\footnotesize OOD Score: \textbf{#2}$^*$}%
    \else
      {\footnotesize OOD Score: \textbf{#2}}%
    \fi
  \end{minipage}%
}
\begin{tabular}{@{} >{\centering\arraybackslash}p{1.05cm} @{\hspace{2pt}} c @{\hspace{2pt}} c @{\hspace{2pt}} c @{}}
\toprule
{\footnotesize\textbf{Patient}} & {\footnotesize\textbf{Visit 1}} & {\footnotesize\textbf{Visit 2}} & {\footnotesize\textbf{Visit 3}} \\
\midrule
\addlinespace[2pt]
{\shortstack{\textbf{Patient 1}\\[2pt]{\footnotesize Healing}}}
&
\casecell{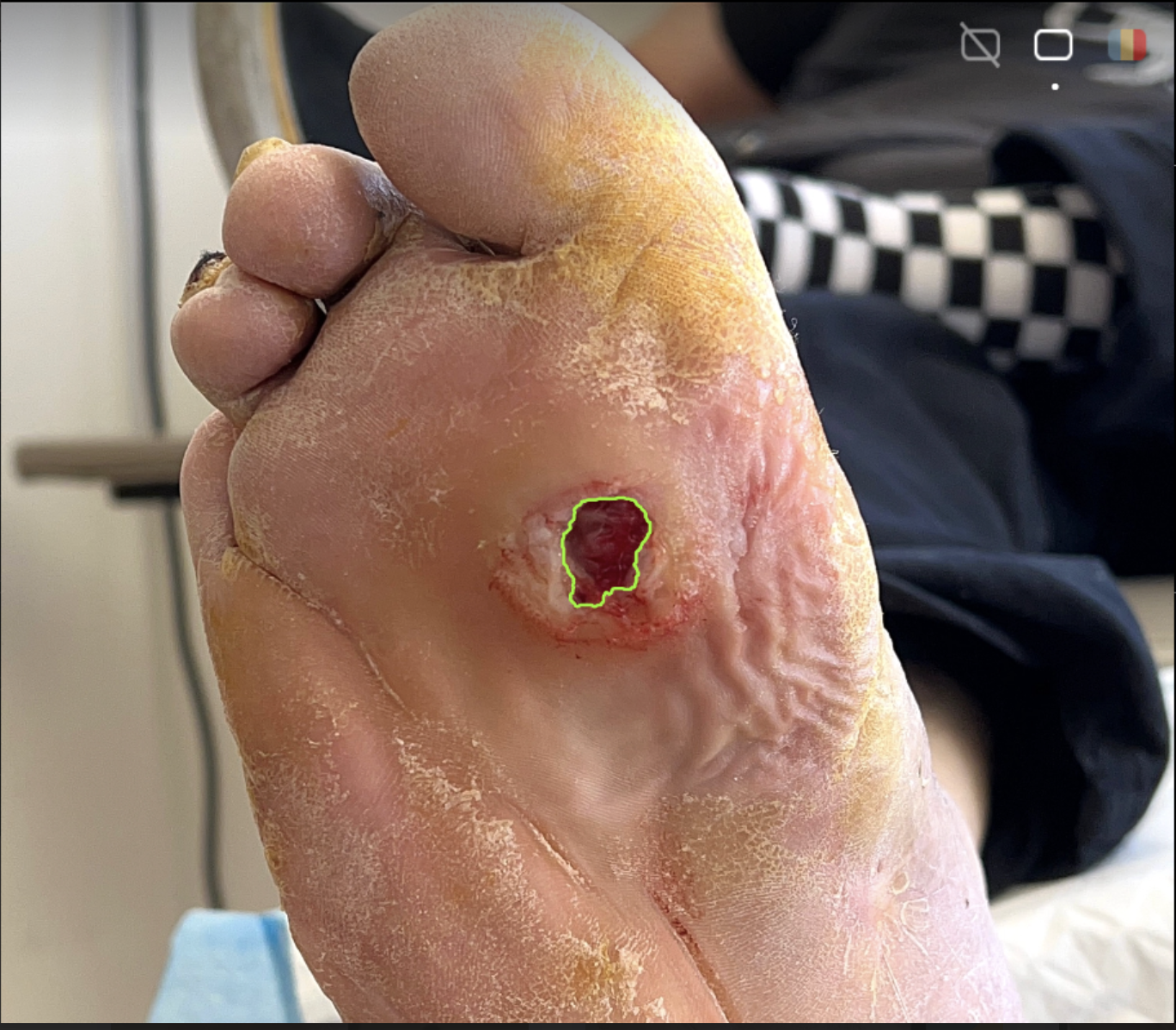}{0.551}{52}{7.3}{2A}%
  {Red tissue with mild dead skin around the wound.}
&
\casecell{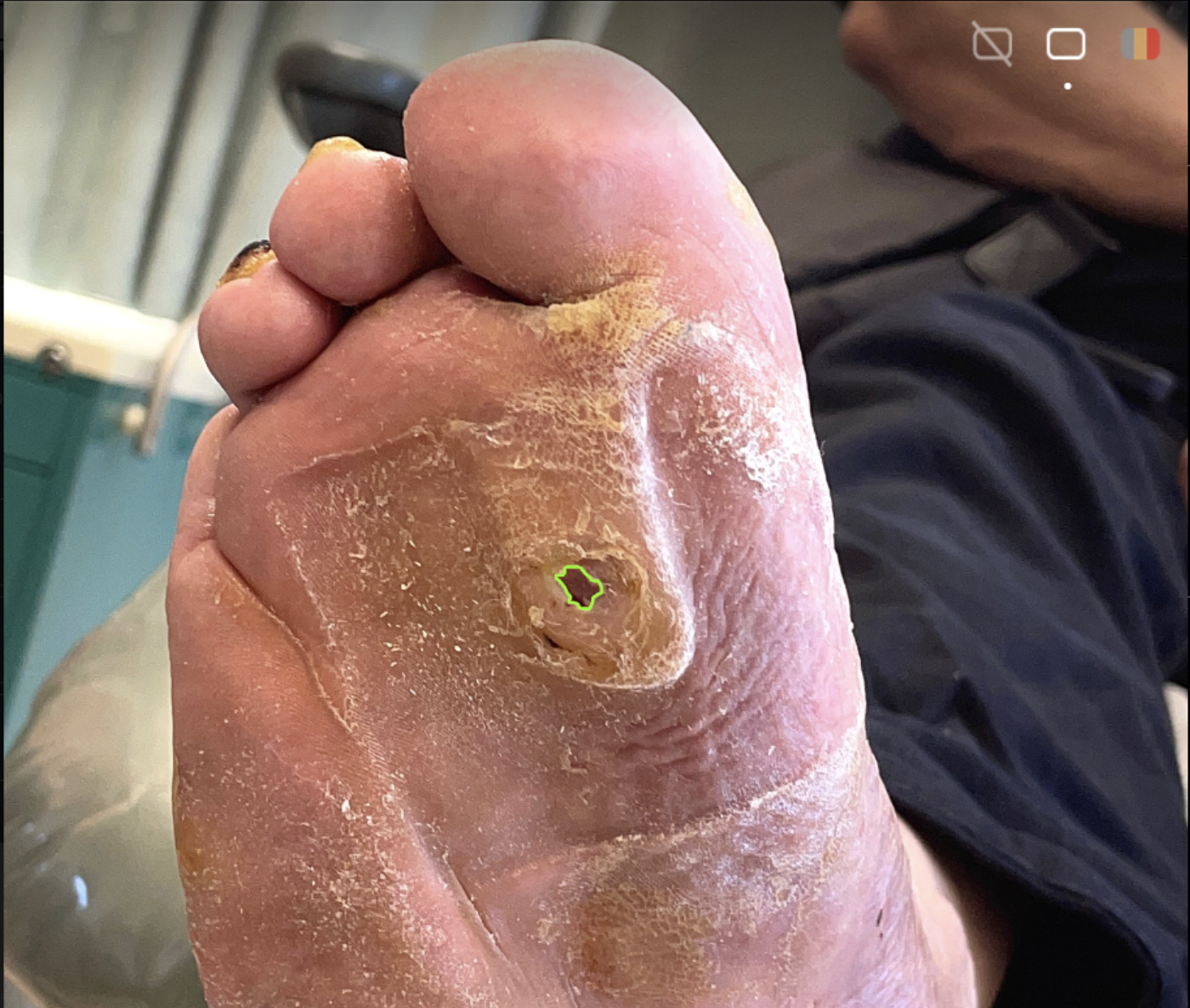}{0.401}{52}{7.3}{2A}%
  {Wound is moist and filling in with new tissue.}
&
\casecell{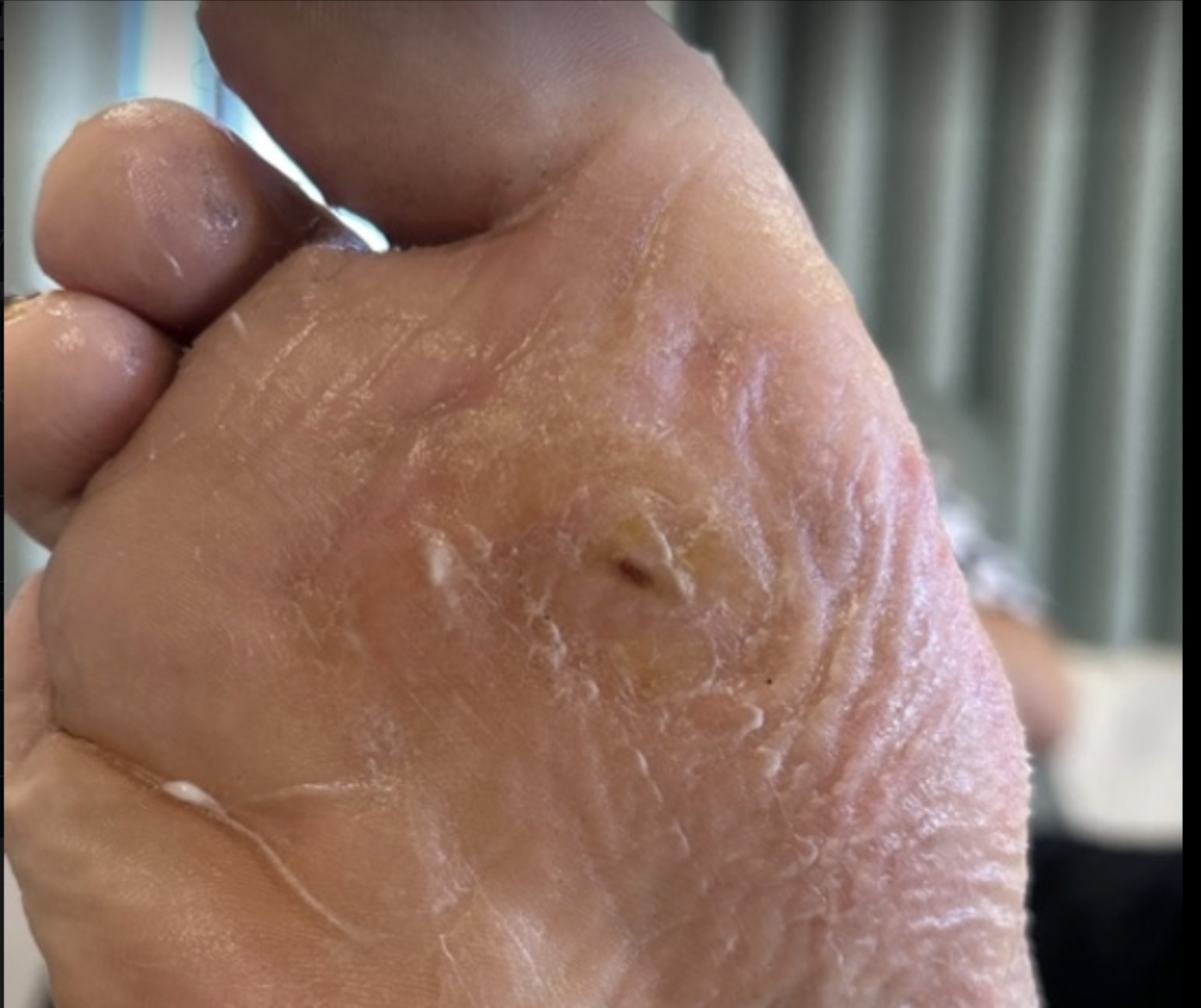}{0.367}{52}{7.3}{2A}%
  {Wound is nearly closed with dry surrounding skin.}
\\
\addlinespace[2pt]
\midrule
\addlinespace[2pt]
{\shortstack{\textbf{Patient 2}\\[2pt]{\footnotesize Non-healed}}}
&
\casecell{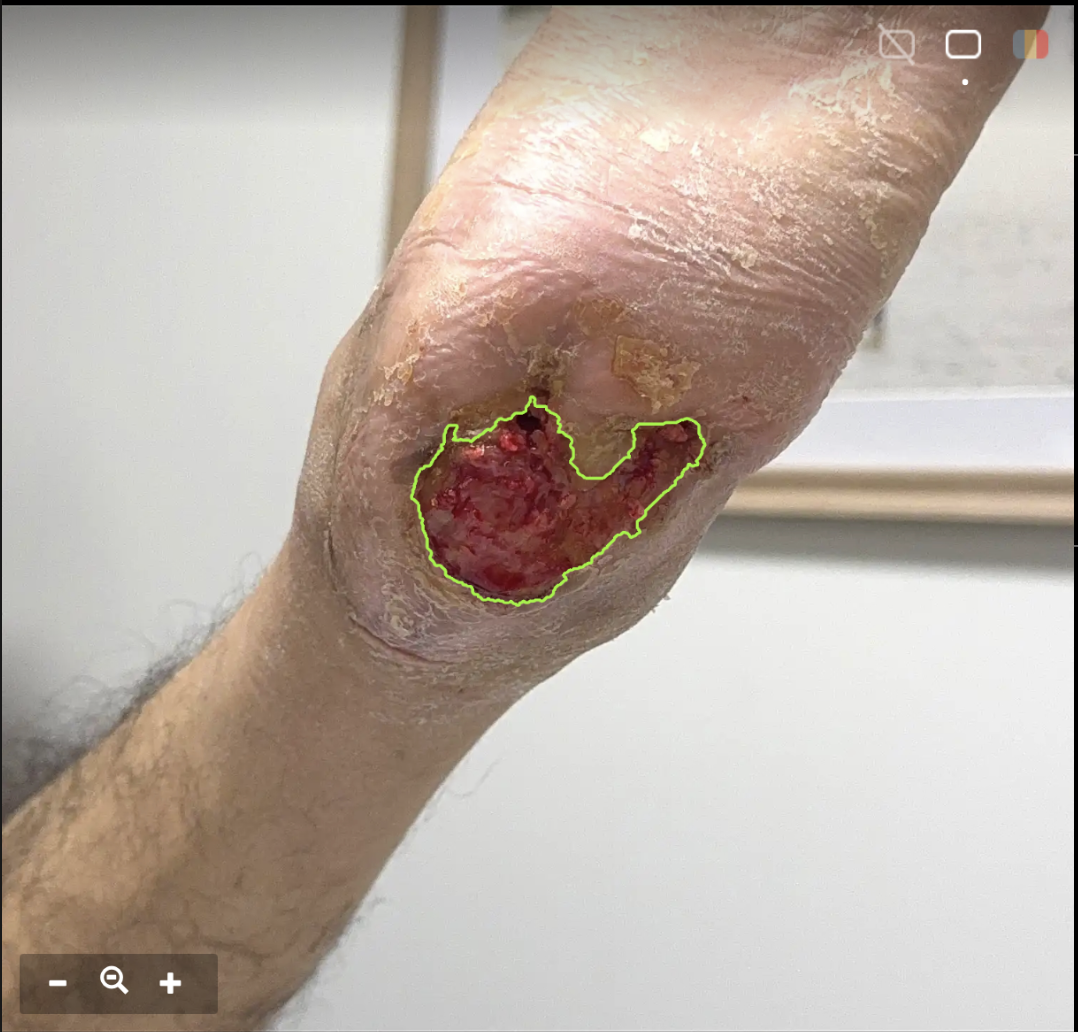}{0.629}{55}{10.0}{1A}%
  {Mixed red and yellow tissue with swollen skin around it.}
&
\casecell{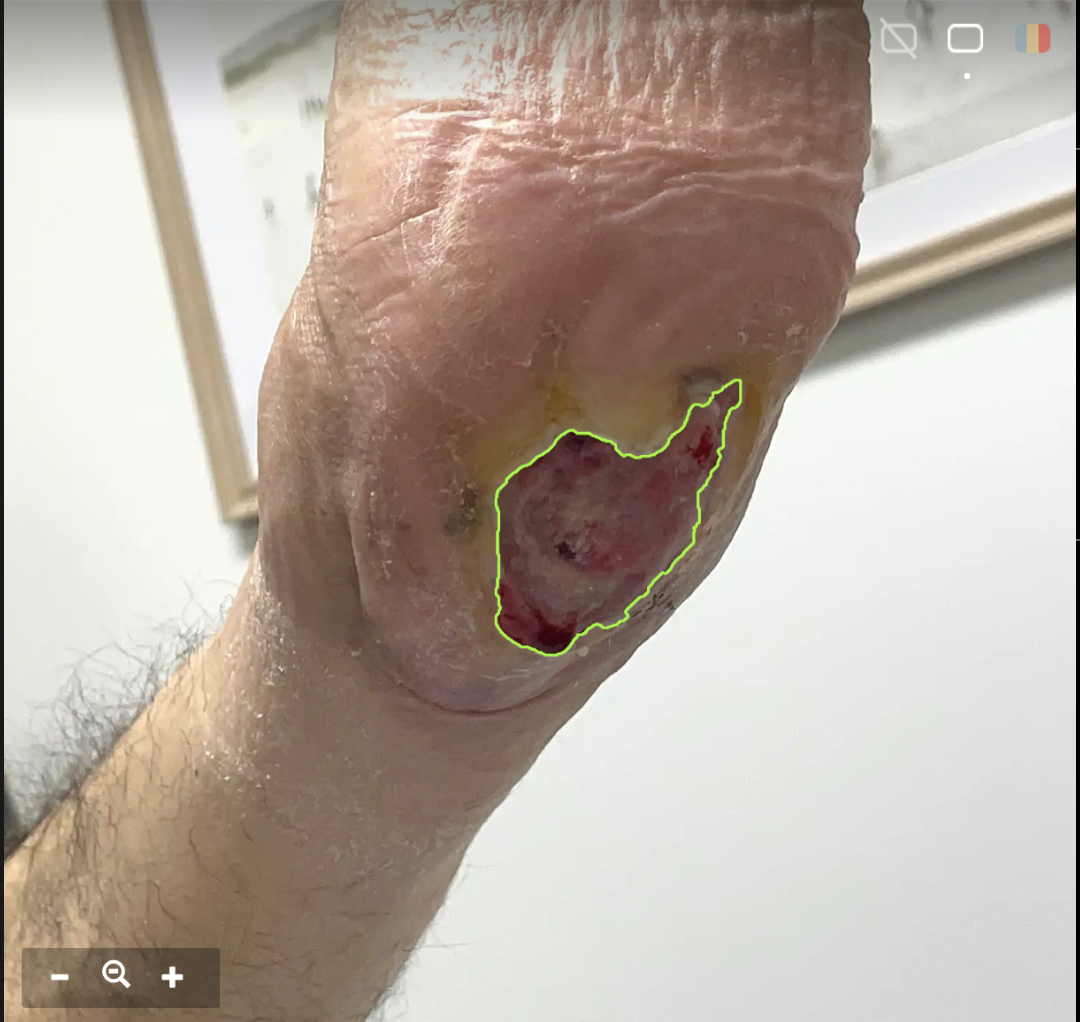}{0.672}{55}{10.0}{1A}%
  {Dark wet tissue with swelling and softened wound edges.}
&
\casecell{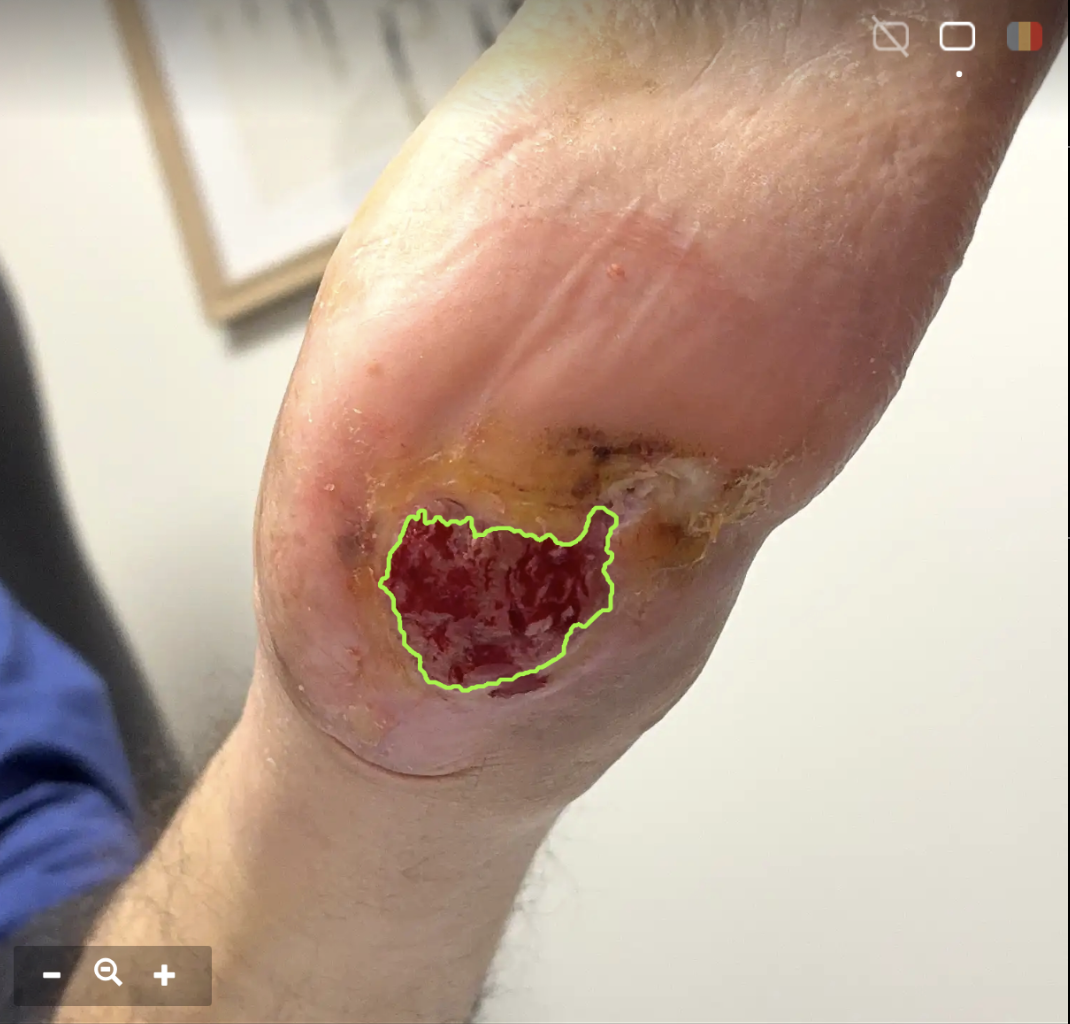}{0.685}{55}{10.0}{1A}%
  {Yellow dead tissue remains with no signs of improvement.}
\\
\addlinespace[2pt]
\midrule
\addlinespace[2pt]
{\shortstack{\textbf{Patient 3}\\[2pt]{\footnotesize AE/SAE}}}
&
\casecell{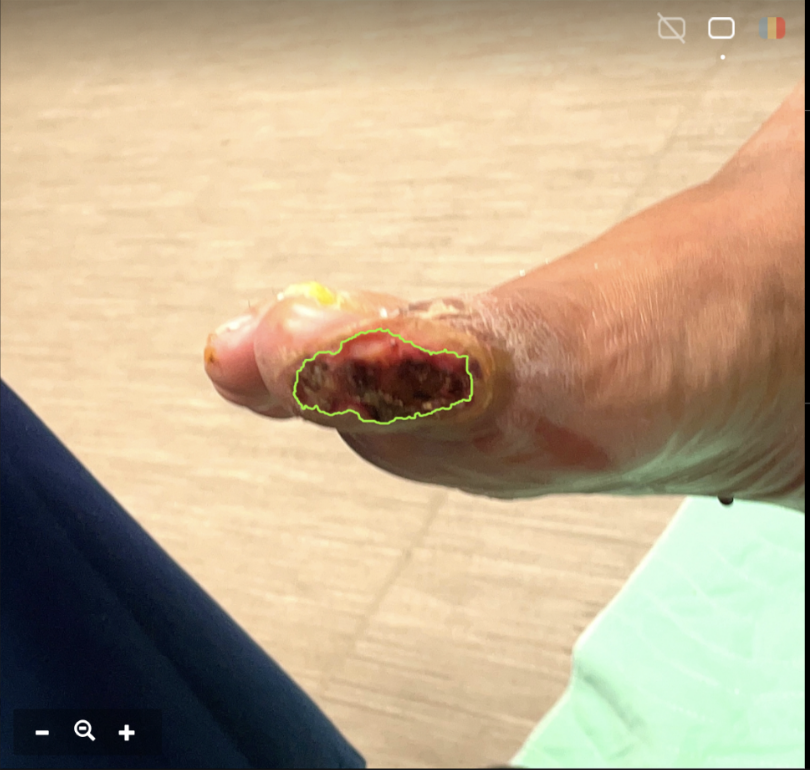}{0.674}{46}{6.7}{2A}%
  {Mixed red and yellow tissue with moderate swelling.}
&
\casecell{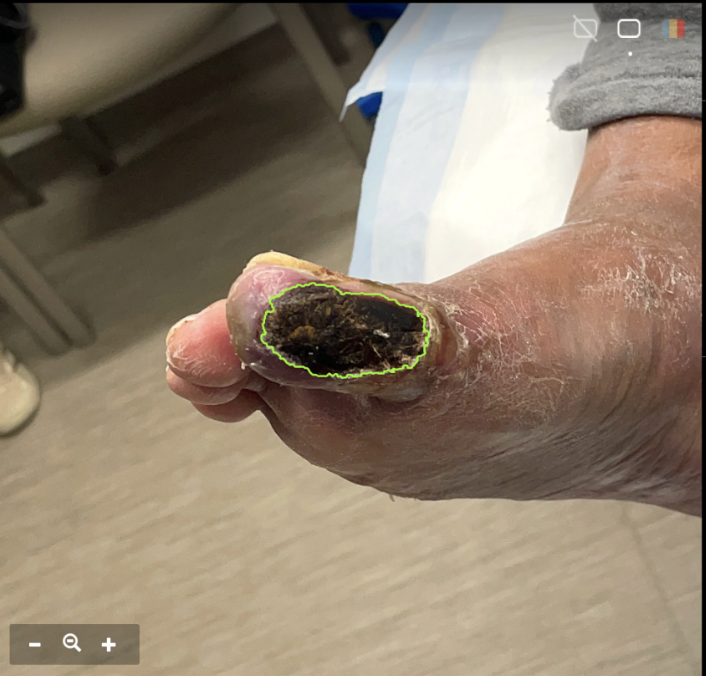}{0.725}{46}{6.7}{2A}%
  {Dead tissue spreading with redness around the wound.}
&
\casecell{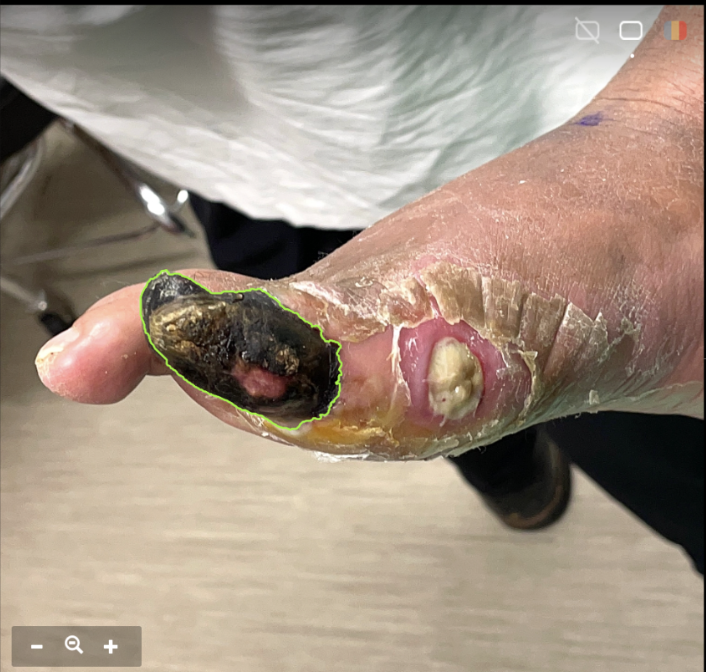}{0.894}{46}{6.7}{2A}%
  {Deep wound cavity with dead tissue and infection signs.$^*$}
\\
\addlinespace[2pt]
\bottomrule
\end{tabular}

\caption{Representative wound trajectories for three patients across 
  longitudinal visits. For each visit, we show the structured clinical 
  context ($T_c$: age, HbA1c, UT grade), a plain-language wound 
  description ($T_d$), and an OOD score measuring how much the sample 
  deviates from the training distribution. Higher OOD scores indicate 
  greater anomaly.}
\label{fig:patient_case_study}
\end{figure*}

\subsection{Threshold Sensitivity Analysis}
The decision threshold $\delta$ is calibrated as the $\delta_q$-th percentile of the training ID score distribution and can be adjusted at inference without retraining, providing a flexible sensitivity--specificity trade-off for different clinical workflows.
Table~\ref{tab:threshold_analysis} sweeps $\delta_q$ from the 50th to the 99th percentile.
Lowering the threshold increases AE/SAE recall at the cost of higher FPR, while raising it reduces false alarms but sacrifices sensitivity.
The default 95th-percentile operating point (recall~0.25, FPR~0.057) is conservative.
Relaxing to the 90th percentile doubles recall to~0.50 while keeping FPR below~0.10, offering an attractive trade-off for screening-oriented deployment.

\begin{table}[t]
\centering
\caption{Effect of different confidence thresholds on recall and false positive rate (FPR) for abnormal wound trajectory detection. The 90\% threshold provides the best trade-off between maintaining recall and minimizing false positives.}
\label{tab:threshold_analysis}
\begin{tabular}{ccc}
\toprule
\textbf{Threshold (\%)} & \textbf{Recall $\uparrow$} & \textbf{FPR $\downarrow$} \\
\midrule
50 & 0.75 & 0.426 \\
60 & 0.50 & 0.312 \\
70 & 0.50 & 0.255 \\
75 & 0.50 & 0.206 \\
80 & 0.50 & 0.191 \\
85 & 0.50 & 0.149 \\
\textbf{90} & \textbf{0.50} & \textbf{0.099} \\
95 & 0.25 & 0.057 \\
98 & 0.00 & 0.021 \\
99 & 0.00 & 0.014 \\
\bottomrule
\end{tabular}
\end{table}

\subsection{Robustness to Covariate Shifts} 

To evaluate robustness under realistic acquisition variations, we apply seven image-level perturbations to the SmartBoot benchmark without altering clinical labels. As shown in Table~\ref{tab:covariate_shift}, the framework maintains stable AUROC (0.504--0.640) and low FPR (0.064--0.128) across all perturbation types, demonstrating that even when visual appearance is degraded, the complementary textual modalities $T_c$ and $T_d$ preserve anomaly detection by supplying context that is immune to image-domain perturbations.

\begin{table}[t]
\centering
\caption{Robustness of the proposed framework under various covariate shifts. Higher AUROC and ID accuracy indicate better performance, whereas lower FPR is preferred. }
\label{tab:covariate_shift}

\begin{tabular}{lccc}
\hline
Covariate Shift & AUROC $\uparrow$ & FPR $\downarrow$ & ID Accuracy $\uparrow$ \\
\hline
Gaussian Blur     & 0.640 & 0.071 & 0.929 \\
Brightness        & 0.626 & 0.106 & 0.894 \\
Contrast          & 0.617 & 0.064 & 0.936 \\
Gaussian Noise    & 0.608 & 0.078 & 0.922 \\
Rotation          & 0.606 & 0.113 & 0.887 \\
Translation       & 0.514 & 0.113 & 0.887 \\
Crop + Resize     & 0.504 & 0.128 & 0.872 \\
\hline
\end{tabular}
\end{table}

\subsection{Ablation Study}

\noindent\textbf{Modality contribution.}
Table~\ref{tab:modality_ablation} shows progressive modality ablation.
Image-only performs worst (AUROC 0.411), confirming visual appearance alone is insufficient.
Adding $T_c$ lifts AUROC to 0.651, substituting $T_d$ instead raises it to 0.674 with a larger FPR95 reduction, as wound descriptions directly encode tissue composition and periwound appearance.
The full trimodal model achieves the strongest results, demonstrating that $T_c$ and $T_d$ provide complementary information.

\begin{table}[t]
\centering
\caption{Modality ablation. \textbf{Bold} = best.}
\label{tab:modality_ablation}
\resizebox{\columnwidth}{!}{%
\begin{tabular}{lccc}
\toprule
Configuration & AUROC\,$\uparrow$ & FPR95\,$\downarrow$ & ID Acc\,$\uparrow$ \\
\midrule
Image Only & 0.411 & 0.709 & 0.857 \\
Image + $T_c$ & 0.651 & 0.695 & 0.859 \\
Image + $T_d$ & 0.674 & 0.651 & 0.915 \\
Image + $T_c$ + $T_d$ (Ours) & \textbf{0.729} & \textbf{0.490} & \textbf{0.937} \\
\bottomrule
\end{tabular}%
}
\end{table}

\noindent\textbf{Cross-contextual fusion.}
Table~\ref{tab:fusion_ablation} isolates each adapter pathway.
Removing either cross-stream term ($\lambda B_c A_d$ or $\mu B_d A_c$) causes the largest AUROC drops (0.548 and 0.524 respectively), confirming bidirectional cross-modal interaction is the primary driver of SAE detection gains. Ablating all cross fusion reduces AUROC to 0.445.

\begin{table}[t]
\centering
\caption{Cross-contextual LoRA fusion ablation. \textbf{Bold} = best.}
\label{tab:fusion_ablation}
\begin{tabular}{lccc}
\toprule
Configuration       & AUROC\,$\uparrow$ & FPR95\,$\downarrow$ & ID Acc\,$\uparrow$ \\
\midrule
Full Model          & \textbf{0.729} & \textbf{0.490} & \textbf{0.937} \\
w/o $\alpha B_cA_c$ & 0.654 & 0.837 & 0.628 \\
w/o $\beta B_dA_d$ & 0.435 & 0.925 & 0.598 \\
w/o $\lambda B_cA_d$& 0.548 & 0.864 & 0.598 \\
w/o $\mu B_dA_c$ & 0.524 & 0.846 & 0.608 \\
w/o Cross Fusion & 0.445 & 0.902 & 0.589 \\
\bottomrule
\end{tabular}
\end{table}

\noindent\textbf{Loss function contribution.}
Table~\ref{tab:loss_ablation} evaluates each objective.
$\mathcal{L}_{\mathrm{COV}}$ reduces the false-positive rate under appearance shifts, and $\mathcal{L}_{\mathrm{TEMP}}$ provides the largest AUROC gain (+0.041 over $\mathcal{L}_{\mathrm{ID}}$ alone) by enforcing cross-visit consistency. 
The full $\mathcal{L}_{\mathrm{TOTAL}}$ achieves
the best results on all metrics.

\begin{table}[t]
\centering
\caption{Loss component ablation. \textbf{Bold} = best.}
\label{tab:loss_ablation}
\begin{tabular}{lccc}
\toprule
Loss Setting & AUROC\,$\uparrow$ & FPR95\,$\downarrow$ & ID Acc\,$\uparrow$ \\
\midrule
$\mathcal{L}_{\mathrm{ID}}$ only                          & 0.663 & 0.627 & 0.891 \\
$\mathcal{L}_{\mathrm{ID}}+\mathcal{L}_{\mathrm{COV}}$   & 0.695 & 0.621 & 0.895 \\
$\mathcal{L}_{\mathrm{ID}}+\mathcal{L}_{\mathrm{TEMP}}$  & 0.704 & 0.572 & 0.902 \\
$\mathcal{L}_{\mathrm{TOTAL}}$                            & \textbf{0.729} & \textbf{0.490} & \textbf{0.937} \\
\bottomrule
\end{tabular}
\end{table}

%% file: sec/5_conclusion.tex
\section{Discussion and Conclusion}
In this paper, we presented a multimodal framework for automated wound monitoring and personalized SAE detection, framing the problem as time-aware OOD detection. Our approach introduces three contributions: a dual-stream cross-contextual LoRA fusion mechanism built on a frozen BiomedCLIP backbone; a wound-specific quadruple cross-modal OOD scoring framework; and an area-reweighted temporal drift penalty that encodes the physiological prior that wound contraction signals recovery. Experiments on the SmartBoot diabetic foot ulcer dataset demonstrate that the proposed framework outperforms existing OOD detection baselines across all evaluated metrics.
The ablation studies surface two broader insights. First, visual appearance alone is a poor basis for SAE detection, confirming that wound images are insufficient without the semantic context clinicians routinely consult. Second, the cross-contextual fusion terms contribute more to OOD performance than the same-stream terms, suggesting that the interaction between clinical context and visual wound descriptors carries diagnostic signal that neither modality encodes independently. Several limitations remain, including the single-trial dataset and the fixed ID text bank, which may limit adaptability as wound descriptions evolve across visits. {\bf Future work} will explore patient-level online adaptation, generalization across diverse wound cohorts, and extension of the temporal OOD formulation to broader post-operative monitoring paradigms. To the best of our knowledge, this is the first framework to address OOD-based SAE detection in clinical wound monitoring.

%% file: sec/Acknowledgement.tex
\section*{Acknowledgements}
Aditi Naiknaware and Salimeh Sekeh have been partially supported by NSF CAREER CCF-2451457. The
findings are those of the authors only and do not represent any position of these
funding bodies.

%% file: sec/x_SM.tex
\clearpage
\label{sec:supplementary}

\section{Extended Related Work}
\label{sec:supp_related}

\noindent\textbf{Wound Image Analysis and DFU Monitoring}

Chronic wound management imposes a substantial global burden: DFUs affect 15--25\% of diabetic patients over their lifetime with 25--44\% one-year recurrence rates following healing~\cite{armstrong2017diabetic,boulton2018}.
A systematic review of image-based AI for wound assessment \cite{anisuzzaman2022review} identifies segmentation, tissue classification, and healing-trajectory prediction as the three central tasks.
Early automated approaches formulated wound care as image segmentation.
Wang et al.~\cite{wang2020woundseg} proposed a lightweight MobileNetV2 framework achieving competitive pixel-level wound delineation.
Scebba et al.~\cite{scebba2022wound} introduced a detect-then-segment pipeline whose use of the eKare inSight 3D platform, the same system as in the SmartBoot trial, yields objective wound-area measurements that feed our temporal drift penalty directly.
Beyond segmentation, tissue-type classification (granulation, slough, eschar, necrosis) has been addressed by ensemble CNN classifiers~\cite{rostami2021ensemble}.
A line of DFUC grand challenges~\cite{cassidy2022dfuc2021} has provided standardised benchmarks for DFU detection and segmentation, driving year-on-year algorithmic progress.
Multimodal systems fusing wound photographs with structured clinical metadata (ulcer grade, HbA1c, offloading adherence) consistently outperform unimodal classifiers~\cite{anisuzzaman2022multimodal}.
The closest prior system, DM-WAT~\cite{cruciani2025dmwat}, fuses DeiT image features with DeBERTa clinical note embeddings via intermediate feature fusion, achieving 77\% referral-decision accuracy.
Our framework differs in three key ways: (i) we use the domain-specific BiomedCLIP backbone, (ii) we generate structured wound descriptions $T^d$ from a clinical VLM, and (iii) we target unsupervised SAE detection, which no prior wound system addresses.

\noindent\textbf{Biomedical Language Models and VLP Backbones}

BiomedCLIP's text encoder is PubMedBERT~\cite{gu2021pubmedbert}, a BERT model pre-trained from scratch on PubMed abstracts and PMC full-text, which substantially outperforms mixed-domain models on biomedical NLP benchmarks.
CLIP~\cite{radford2021clip} established contrastive image-text pre-training as the paradigm for zero-shot visual recognition via an InfoNCE objective over large web corpora.
Medical adaptations proliferated: MedCLIP~\cite{wang2022medclip} decouples contrastive pairs on chest X-rays with a semantic matching loss to handle false negatives; BioViL-T~\cite{bannur2023biovilt} trains a CNN-Transformer hybrid on prior--current image-report pairs achieving state-of-the-art disease-progression classification. 
BiomedCLIP~\cite{zhang2023biomedclip} scales this to 15M PMC figure-caption pairs.
Large multimodal models extend this further: BLIP-2~\cite{li2023blip2} bridges frozen image encoders to frozen language models via a lightweight Q-Former, LLaVA-Med~\cite{li2024llava_med} instruction-tunes a large VLM for biomedical VQA in a single training day, Med-Flamingo~\cite{moor2023medflamingo} achieves few-shot medical reasoning with a frozen vision backbone.

\noindent\textbf{Parameter-Efficient Fine-Tuning}

Adapter modules~\cite{houlsby2019adapter} were the first parameter-efficient approach, inserting small bottleneck layers into frozen Transformer blocks.
Prefix-tuning~\cite{li2021prefix} prepends learnable continuous vectors to the input and has been shown to match full fine-tuning on generation tasks.
Prompt tuning (CoOp~\cite{zhou2022coop}, CoCoOp~\cite{zhou2022cocoop}) optimises discrete or continuous context tokens for vision-language transfer.
LoRA~\cite{hu2022lora} parameterises weight updates as $\Delta W\!=\!BA$ with $r\!\ll\!\min(d,k)$, reducing trainable parameters to below 1\% while preserving pre-trained behaviour through zero initialisation of $B$.
AdaLoRA~\cite{hu2023adalora} extends LoRA by dynamically allocating the rank budget across weight matrices via singular value decomposition, improving parameter efficiency on tasks where different layers warrant different adaptation capacity.
In medical imaging, LoRA has been applied to radiology report generation (Med-Flamingo~\cite{moor2023medflamingo}), clinical dialogue summarisation, VQA (LLaVA-Med~\cite{li2024llava_med}), and zero-shot BiomedCLIP transfer.
Contrastive self-supervised anomaly pre-training (CSI~\cite{tack2021csi}) provides a complementary direction, showing that distributionally shifted augmentations yield OOD-sensitive representations without labels.
Our contribution extends LoRA to a \emph{dual-stream, cross-contextual} design: separate adapters for clinical context $T^c$ and wound-image descriptions $T^d$ are composed cross-stream as $B_d A_c$ and $B_c A_d$, grounded in~\cite{sekeh2026crossmodal}, and, to our knowledge, is the first LoRA formulation coupling two semantically distinct text streams.

\noindent\textbf{OOD Detection: Classical and Generative Methods}

Classical unimodal OOD detection defines ID membership through confidence.
MSP~\cite{hendrycks2017msp} uses the maximum softmax probability, ODIN~\cite{liang2018odin} improves separation via temperature scaling and input perturbations,
Energy~\cite{liu2020energy} replaces softmax with a theoretically grounded free-energy score, Outlier Exposure~\cite{hendrycks2019deep} trains on auxiliary negatives to push OOD posterior distributions toward uniform.
Activation-shaping approaches, such as ReAct~\cite{sun2021react} and ASH~\cite{djurisic2023ash}, rectify or ablate penultimate-layer activations to suppress OOD overconfidence.
ViM~\cite{zhang2022vim} projects residual feature magnitudes into the null space of the classifier to produce an energy-free OOD score.
Distance-based methods, like Mahalanobis~\cite{lee2018mahalanobis} and deep kNN~\cite{sun2022knnood}, score inputs by proximity to class-conditional feature-space prototypes or neighbourhoods.
Generative approaches have complemented these: autoencoders for brain MRI anomaly detection~\cite{baur2021autoencoders} and, more recently, denoising diffusion models for unsupervised OOD
detection~\cite{graham2023diffusion} demonstrate that reconstruction error from generative models serves as a reliable anomaly signal across diverse imaging modalities.
The OpenOOD benchmark~\cite{yang2022openood} and the accompanying generalized survey~\cite{yang2024generalized} provide unified evaluation protocols that encompass anomaly detection, open-set recognition, and distributional shift under a common framework.

\noindent\textbf{OOD Detection with Vision-Language Models}

Vision-language models introduced a multi-modal OOD detection paradigm.
MCM~\cite{ming2022ood} was the first to leverage CLIP's aligned embeddings for zero-shot detection, scoring images by maximum temperature-scaled cosine similarity to ID class-name text embeddings, outperforming Mahalanobis by $>$13\% AUROC on semantically hard tasks.
GL-MCM~\cite{miyai2025glmcm} extends MCM with local patch scores for multi-object scenes, CLIPN~\cite{wang2023clipn} adds learned ``no'' prompts for threshold-free inference, LoCoOp~\cite{miyai2023locoop} combines few-shot prompt tuning with local OOD awareness, and NegLabel~\cite{jiang2024neglabel} selects discriminative negative class names from large vocabularies as an explicit OOD reference.
WinCLIP~\cite{jeong2023winclip} adapts CLIP for zero- and few-shot anomaly detection in industrial images via compositional prompt ensembles and multi-scale window-based feature matching, a design related to our data-driven ID text bank construction from fused LoRA embeddings.
For medical imaging specifically, Hong et al.~\cite{hong2024medood} identify unsupervised label-free scenarios (our SAE setting) as the key open problem.
T-QPM~\cite{naiknaware2026tqpm} unifies four cross-modal score signals with a temporal drift penalty.
We adapt it to wound care by deriving ID prototypes from cross-contextual LoRA embeddings and reweighting the temporal penalty by sigmoid-normalised wound-area change (grounded theoretically by~\cite{sekeh2026crossmodal}).

\noindent\textbf{Temporal Modeling in Clinical AI}

Temporal dynamics are a defining characteristic of wound care: area contraction signals healing, while stagnation or growth signals adverse risk.
Self-attention mechanisms~\cite{vaswani2017attention} have become the dominant architecture for modeling long-range temporal dependencies in sequential data, with extensive adoption in clinical time-series.
BioViL-T~\cite{bannur2023biovilt} demonstrated that incorporating prior visit images and reports into temporal contrastive objectives substantially improves disease-progression classification in radiology. Autoencoders~\cite{baur2021autoencoders} have been applied to
longitudinal brain MRI anomaly detection, learning expected morphological trajectories and flagging deviations.
Contrastive methods (CSI~\cite{tack2021csi}) learn representations that remain stable under in-distribution covariate shift yet are sensitive to out-of-distribution novelty, properties directly exploited in our covariate consistency loss.
The generalized OOD survey~\cite{yang2024generalized} and the medical OOD review~\cite{hong2024medood} both highlight that models deployed under temporal data drift degrade silently, motivating explicit drift-aware regularisation.
Our area-reweighted penalty $\mathcal{L}_\text{TEMP}$ operationalises this insight for wound physiology: sigmoid-normalised wound-area change between visits reweights the temporal consistency penalty, penalising score inconsistency proportionally to the magnitude of the physiological anomaly signal.
The OpenOOD benchmark~\cite{yang2022openood} also serves as a reference for evaluating temporal robustness under covariate shift, and we use its FPR@TPR95 and AUROC metrics throughout our evaluation.

\section{Details on Quadruple Cross-Modal Scores}
\noindent$S_{\mathrm{ID}}$: Semantic Matching Score 
(image $\leftrightarrow$ ID text bank).
\begin{equation}
  S_{\mathrm{ID}}(I_i)
  = \max_{k \in \{0,1\}} \frac{\mathbf{z}_{\mathrm{ID}}(I_i)[k]}{\tau}.
  \label{eq:s_id}
\end{equation}
\noindent A high value indicates that the wound image is visually consistent with
a known ID healing class; a low value suggests the visual appearance
departs from both normal and moderate healing which could be a potential SAE.

\noindent$S_{\mathrm{VIS}}$: Wound-based Score 
(image $\leftrightarrow$ ID visual prototypes).
\begin{align}
  \mathrm{KL}_k(I_i,t)
  &= \sum_{j=0}^{1} p_j(I_i)\log\frac{p_j(I_i)}{\mu_{k,t}[j]},\nonumber\\
  S_{\mathrm{VIS}}(I_i,t)
  &= -\min_{k \in \{0,1\}}\;\mathrm{KL}_k(I_i,t).
  \label{eq:s_vis}
\end{align}
\noindent
A higher $S_{\mathrm{VIS}}$ indicates the wound's visual probability
pattern conforms to expected ID behaviour at post-operative day~$t$.

\noindent $S_{\mathrm{CAP\text{-}T}}$: Caption-Text Alignment Score (fused caption $\leftrightarrow$ ID text bank): $  S_{\mathrm{CAP\text{-}T}}(I_i) = \max_{k \in \{0,1\}}\; \langle \mathbf{q}_i,\, \mathbf{t}_k \rangle.$

\noindent
Because $\mathbf{q}_i$ and $\mathbf{t}_k$ both originate from the same
LoRA-adapted encoder and projection head, this score measures how closely
the test sample's joint clinical-visual semantics align with the aggregated
semantics of known ID classes,  a comparison that is both domain-adapted
and directly meaningful.

\noindent $S_{\mathrm{CAP\text{-}V}}$: Caption-Visual Alignment Score (fused caption $\leftrightarrow$ ID visual prototypes). This score is computed by
\begin{equation*}
  S_{\mathrm{CAP\text{-}V}}(I_i,t)
  = -\min_{k \in \{0,1\}}\;\mathrm{KL}\!\left(
      \mathbf{p}_{\mathrm{CAP}}(I_i)\;\Big\|\;\boldsymbol{\mu}_{k,t}
    \right),\;\hbox{where},
  \label{eq:s_capv}
\end{equation*}
\begin{align*}
  \mathbf{z}_{\mathrm{CAP}}(I_i)
  &= \bigl[\mathbf{q}_i^\top\mathbf{t}_0,\;
           \mathbf{q}_i^\top\mathbf{t}_1\bigr]^\top,\\
  \mathbf{p}_{\mathrm{CAP}}(I_i)
  &= \mathrm{Softmax}\!\!\left(\frac{\mathbf{z}_{\mathrm{CAP}}(I_i)}{\tau}\right),
  \label{eq:p_cap}
\end{align*}
\section{Implementation Details}

All experiments were implemented in PyTorch and conducted on a single NVIDIA L40 GPU. The proposed framework utilizes BiomedCLIP as the multimodal vision-language backbone, where wound images are encoded using the pretrained visual encoder and clinical text is processed through the pretrained text encoder. To efficiently adapt the foundation model to the wound OOD detection task, Low-Rank Adaptation (LoRA) was employed while keeping the original BiomedCLIP parameters frozen.

All wound images were resized to $224\times224$ pixels and normalized using the standard BiomedCLIP preprocessing pipeline. Structured clinical information ($T_c$) and automatically generated wound descriptions ($T_d$) were tokenized using the BiomedCLIP tokenizer with a maximum context length of 77 tokens.

LoRA adapters were inserted into the text encoder using a rank of $r=4$, scaling factor $\alpha=8$, and dropout rate of $0.1$. During training, only the LoRA parameters together with the proposed multimodal OOD scoring module were optimized, resulting in a parameter-efficient fine-tuning strategy.

Models were trained for 10 epochs using the AdamW optimizer with a learning rate of $3\times10^{-4}$ and a batch size of 16. The temperature parameter used in the multimodal similarity computation was fixed to $\tau=0.07$, while the spatial attention mixing weight was set to $\gamma_s=0.5$. The covariate consistency and temporal consistency objectives were weighted by $\lambda_{cov}=0.1$ and $\lambda_{temp}=0.05$, respectively.

 OOD decisions were obtained using the proposed fused anomaly score, which combines image similarity, multimodal text consistency, and temporal representations. The decision threshold was estimated from the training ID score distribution using percentile-based calibration.

\section{Clinical Description of Wound Dataset}
\label{sec:sm_dataset}

\noindent\textbf{Temporal wound organization.}
For each participant, wound images were chronologically ordered according to the clinical visit sequence extracted from image filenames and study metadata.
Baseline images were assigned to week~0, while subsequent visits were indexed by their corresponding clinical follow-up week.
Images with incomplete metadata, duplicated administrative screenshots, or corrupted acquisitions were removed during preprocessing.

\noindent\textbf{Clinical outcome labeling.}
Patients were categorized into healed, non-healed, and adverse event/severe adverse event (AE/SAE) groups using the final clinical outcome recorded in the SmartBoot database.
All images from healed and non-healed patients were treated as
in-distribution (ID) samples.
For patients who experienced an AE or SAE, the terminal wound image was assigned the OOD label, while earlier observations were retained as ID samples.
This formulation allows the model to learn normal wound evolution from longitudinal observations while reserving the clinically abnormal terminal state as the OOD target.

\noindent\textbf{Multimodal representation.}
Each wound image was paired with two complementary textual modalities.
The first modality ($T_c$) consists of structured clinical information extracted from the SmartBoot database, including demographic information, wound characteristics, ulcer duration and location, glycemic status, and additional structured clinical descriptors available at the corresponding visit.
The second modality ($T_d$) is an automatically generated wound description produced by a large vision-language model.
To avoid leakage of quantitative wound measurements, the prompting strategy explicitly instructed the model to describe only localized wound appearance, tissue appearance, periwound characteristics, visible inflammation, granulation tissue, slough, and necrotic tissue, while ignoring overlaid measurements, interface elements, timestamps, and other non-clinical annotations present in the eKare images.

\noindent\textbf{Patient-level data partitioning.}
To prevent patient-specific information leakage across splits, all experiments used patient-level partitioning rather than image-level partitioning.
Entire patient trajectories were assigned exclusively to training, validation, or testing.
AE/SAE patients were excluded from training and distributed only between validation and testing, ensuring the framework learns exclusively from in-distribution wound trajectories during optimization and is evaluated on previously unseen abnormal wound progressions during inference.

\begin{figure*}[t]
\centering
\scriptsize
\setlength{\tabcolsep}{4pt}
\renewcommand{\arraystretch}{0.72}

\newcommand{\casecell}[6]{%
  \begin{minipage}[t]{0.32\linewidth}
    \centering
    \includegraphics[width=\linewidth,height=1.85cm,keepaspectratio]{#1}\\[1pt]
    {\footnotesize $T_c$: Age:~#2, HbA1c:~#3, UT:~#4}\\[1pt]
    {\footnotesize $T_d$: #5}\\[1pt]
    {\footnotesize OOD Score: #6}
  \end{minipage}%
}

\newcommand{\emptycell}{%
  \begin{minipage}[t]{0.32\linewidth}
    \centering
    {\footnotesize --}
  \end{minipage}%
}

\begin{tabular}{@{} >{\centering\arraybackslash}p{1.05cm} @{\hspace{2pt}} c @{\hspace{2pt}} c @{\hspace{2pt}} c @{}}
\toprule
{\footnotesize\textbf{Patient}} &
{\footnotesize\textbf{Visit 1}} &
{\footnotesize\textbf{Visit 2}} &
{\footnotesize\textbf{Visit 3}} \\
\midrule


{\shortstack{\textbf{Patient 4}\\[2pt]{\footnotesize Healed}}}
&
\casecell
{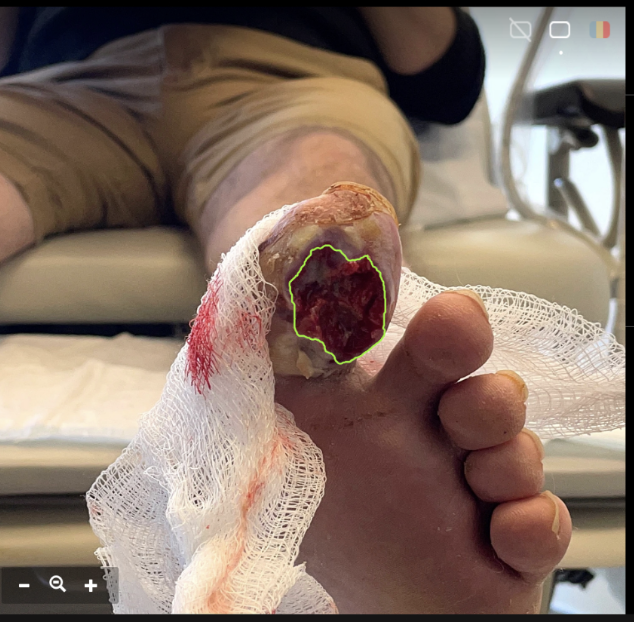}
{34}{12.0}{2A}
{Dark red granulation with fibrin/slough.}
{0.681}
&
\casecell
{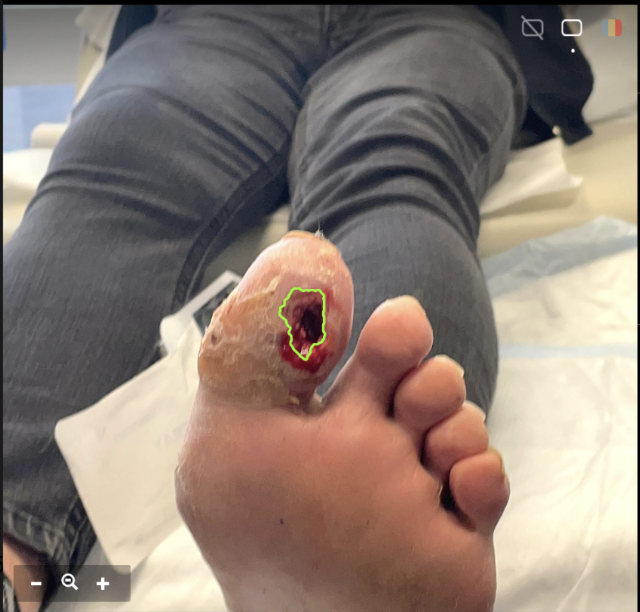}
{34}{12.0}{2A}
{Deep red granulation with maroon center.}
{0.556}
&
\casecell
{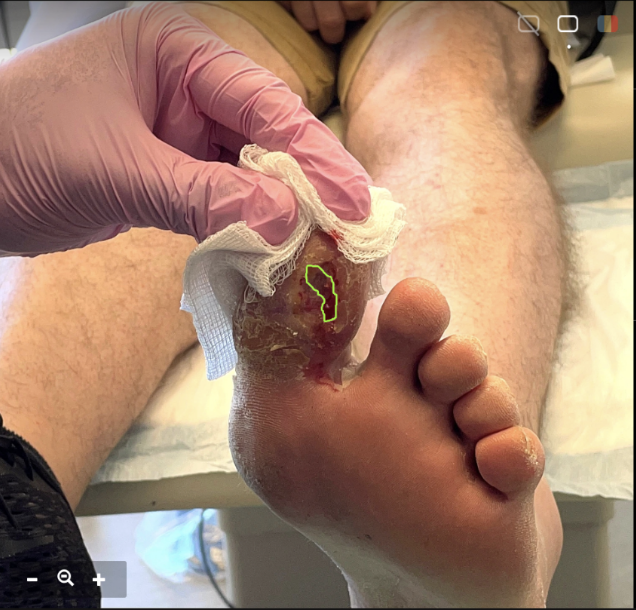}
{34}{12.0}{2A}
{Dark moist granulation tissue.}
{0.349}
\\

\midrule


{\shortstack{\textbf{Patient 5}\\[2pt]{\footnotesize Non-Healed}}}
&
\casecell
{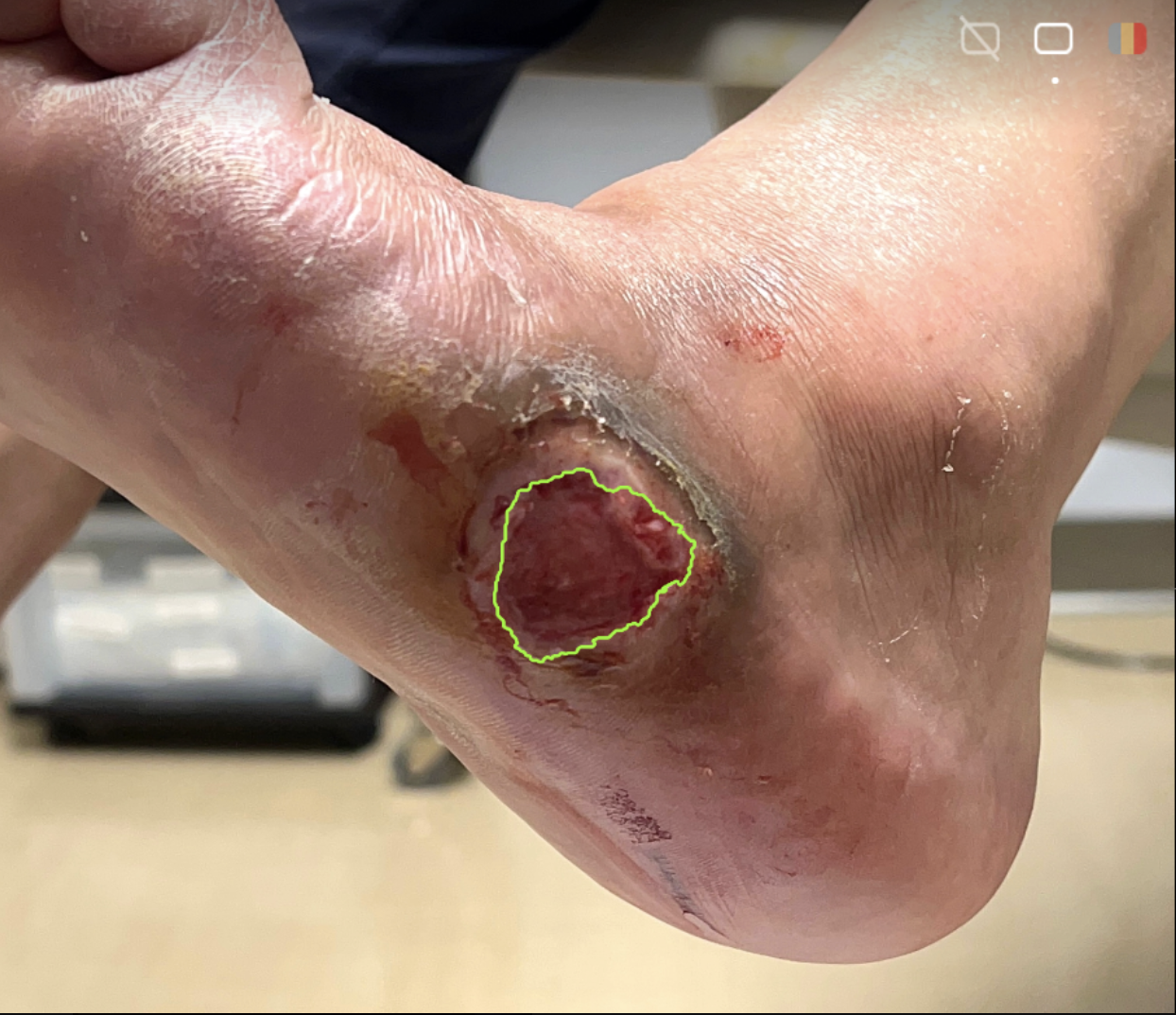}
{51}{4.5}{2A}
{Large red granulation wound with mild slough.}
{0.491}
&
\casecell
{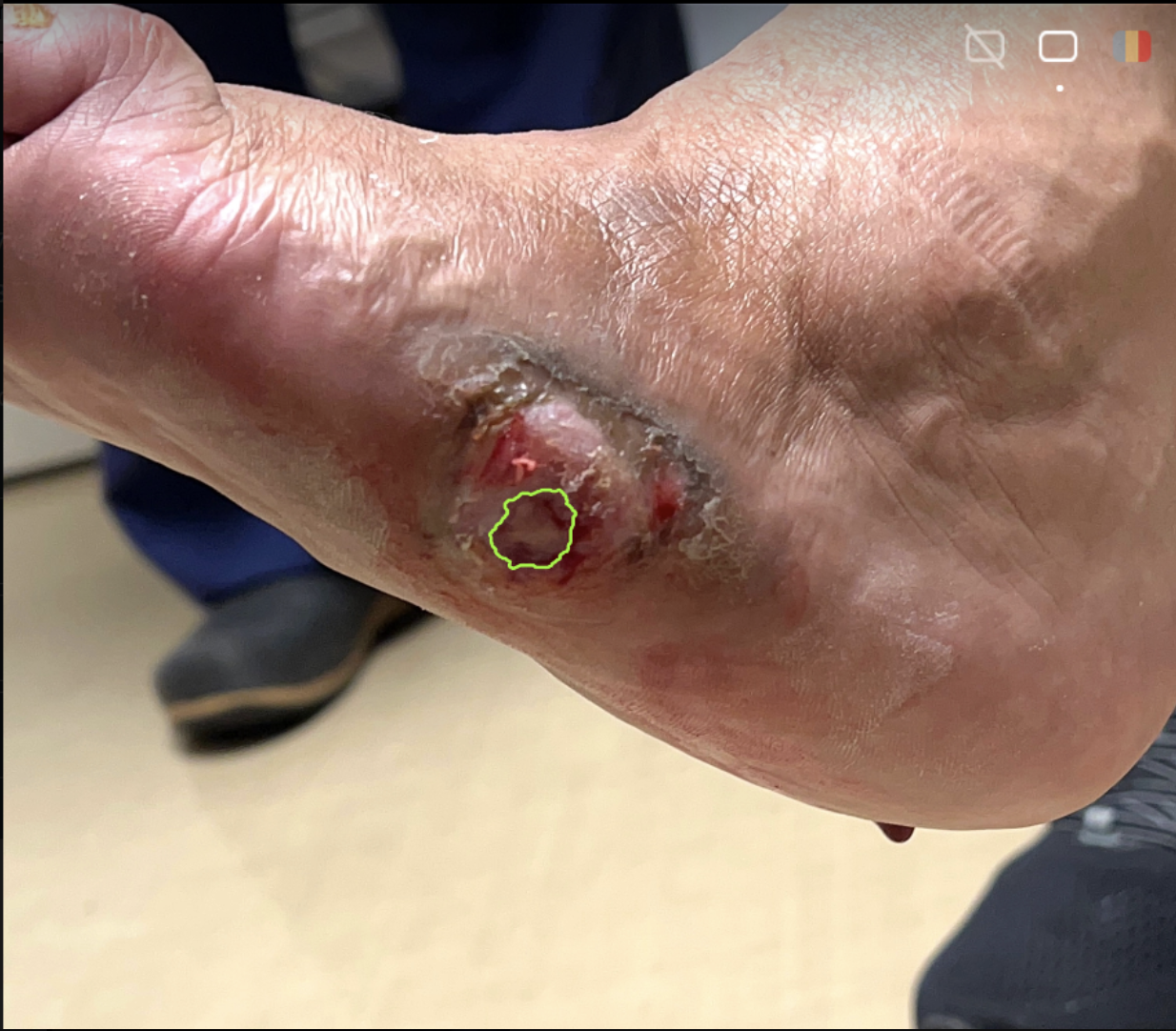}
{51}{4.5}{2A}
{Smaller wound with central dark tissue and healing margins.}
{0.408}
&
\casecell
{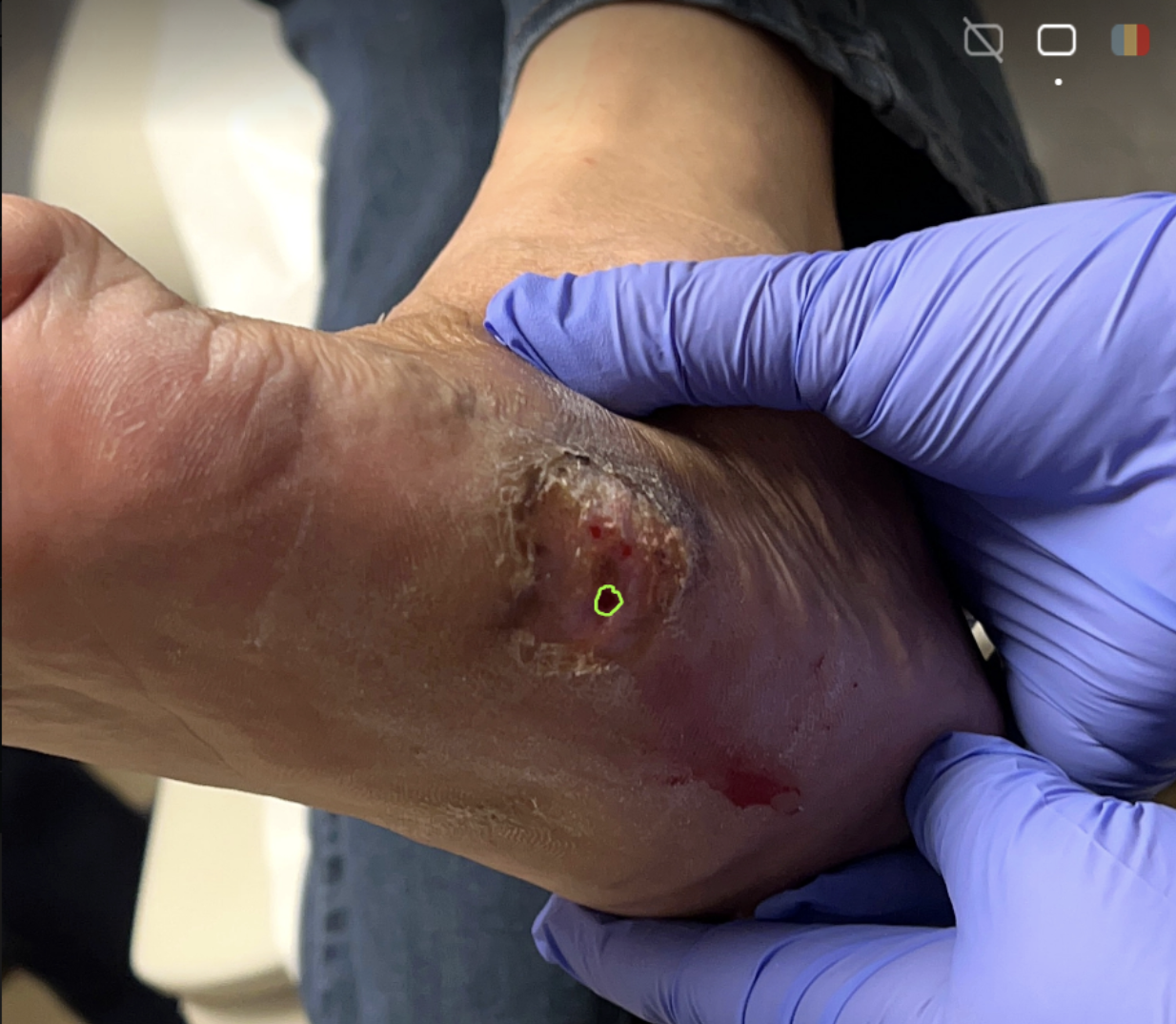}
{51}{4.5}{2A}
{Tiny residual ulcer with epithelialization; nearly healed.}
{0.396}
\\

\midrule


{\shortstack{\textbf{Patient 6}\\[2pt]{\footnotesize AE/SAE}}}
&
\casecell
{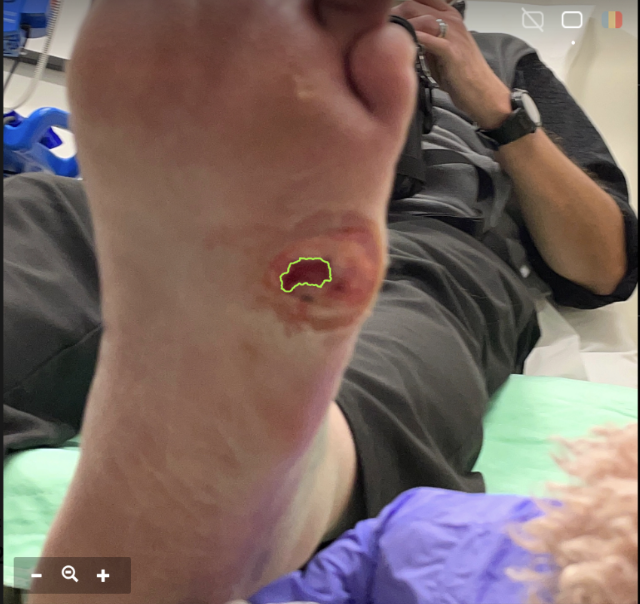}
{57}{8.2}{1A}
{Bright red moist granulation.}
{0.364}
&
\casecell
{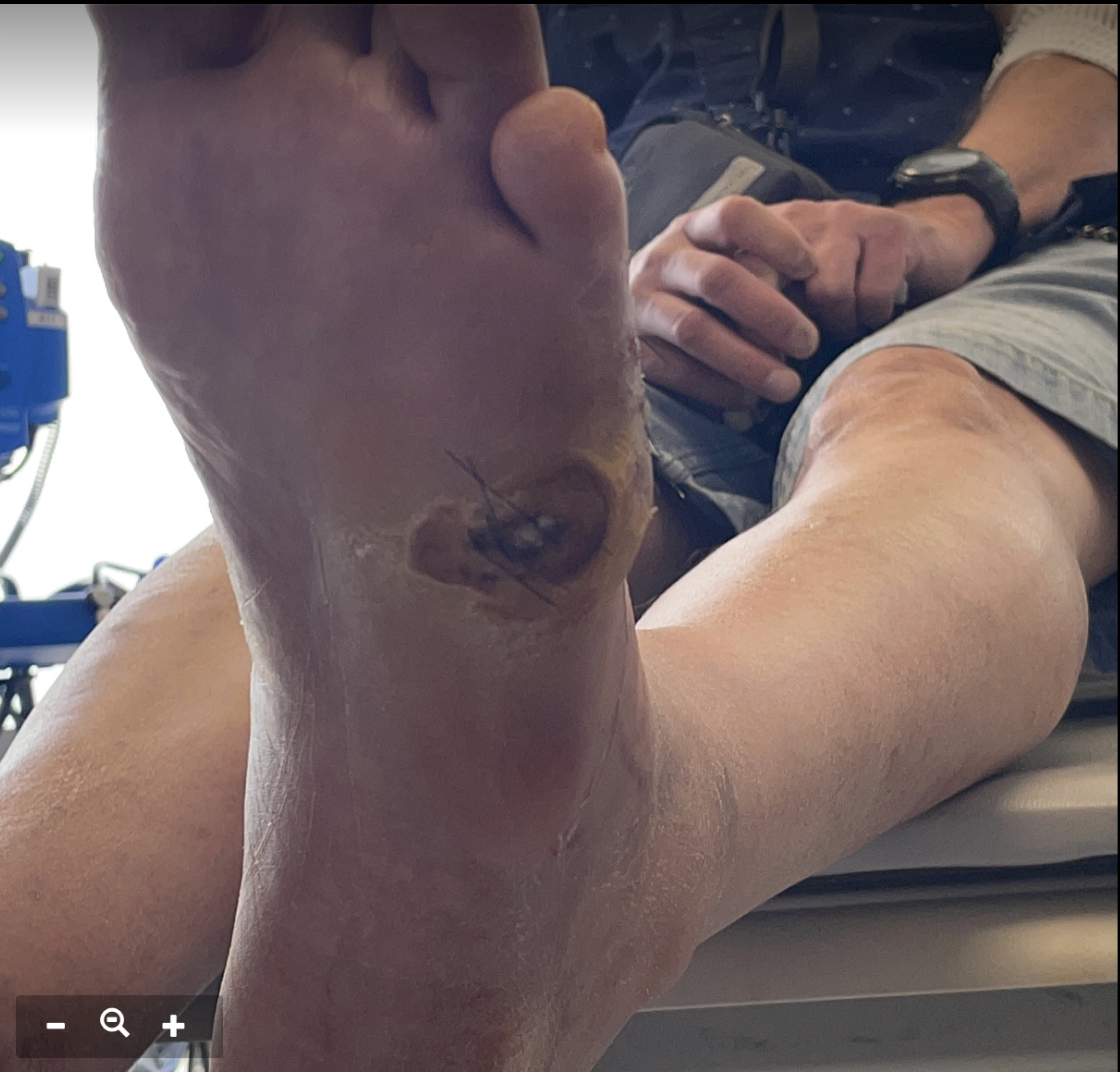}
{57}{8.2}{1A}
{Mixed necrotic and hemorrhagic tissue.}
{0.552}
&
\emptycell
\\

\midrule


{\shortstack{\textbf{Patient 7}\\[2pt]{\footnotesize AE/SAE}}}
&
\casecell
{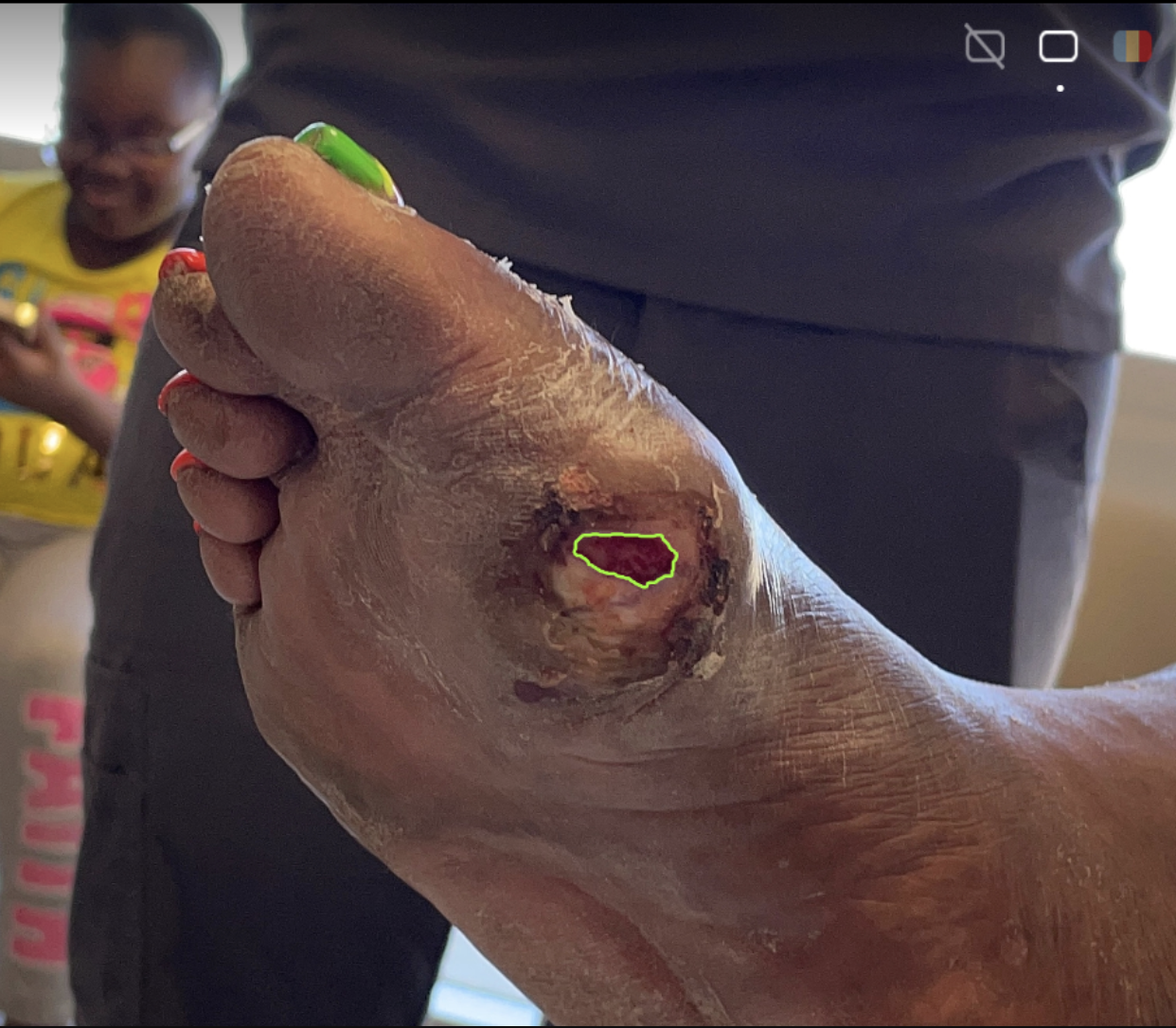}
{43}{6.2}{1A}
{Red granulation with darker tissue.}
{0.562}
&
\casecell
{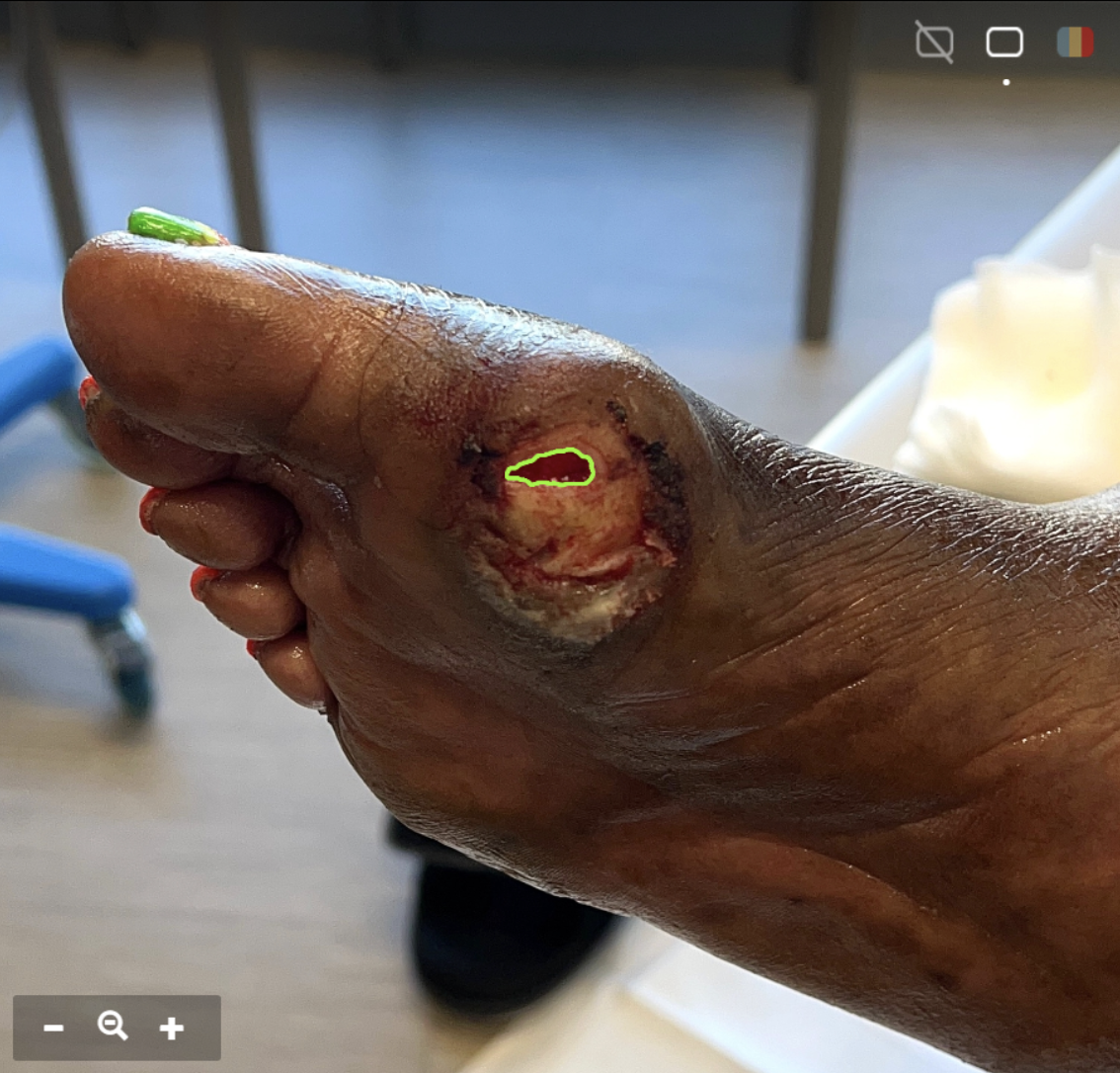}
{43}{6.2}{1A}
{Red granulation mixed with yellow slough.}
{0.593}
&
\emptycell
\\

\bottomrule
\end{tabular}

\caption{Representative wound trajectories for five patients across longitudinal visits. For each visit, we show the structured clinical context ($T_c$: age, HbA1c, UT grade), a brief wound description ($T_d$), and the corresponding OOD score produced by our model. Higher OOD scores indicate greater deviation from the training distribution and potentially increased risk of adverse outcomes.}
\label{fig:patient_case_study}
\end{figure*}
\section{Training Dynamics}
\label{sec:sm_dynamics}

Figure 3 in the main paper shows the three individual loss components over 10 epochs.
$\mathcal{L}_{\mathrm{ID}}$ decreases rapidly in early epochs, reflecting stable adaptation to in-distribution wound categories.
$\mathcal{L}_{\mathrm{COV}}$ remains consistently low throughout, confirming invariant OOD anomaly scoring under visual perturbations.
$\mathcal{L}_{\mathrm{TEMP}}$ decreases steadily, stabilizing longitudinal score representations across successive clinical visits.
The simultaneous minimization across all three objectives demonstrates robust multimodal optimization using parameter-efficient LoRA adapters, with no noticeable oscillations or divergence across the 10 training epochs.

Validation AUROC begins at $\approx$0.43 (epoch~1), rises to a peak of 0.75--0.76 (epochs 8--9), then settles at $\approx$0.69 at the final epoch.
Despite minor mid-training fluctuations, which is expected given the small-sample OOD setting, the stable convergence region confirms the model does not overfit severely despite the limited number of AE/SAE patients.
FPR95 falls from $\approx$0.93 (epoch~1) to $\approx$0.41 (epoch~9), a reduction exceeding 50\%, before rising slightly to 0.49 at the final epoch. 
The overall downward trend demonstrates that the fused anomaly score progressively reduces false alarms while maintaining detection capability.
ID accuracy improves from $\approx$63\% (epoch~1) to $\approx$97\% by the final epoch, with only a minor temporary dip around epoch~8 before recovering.
The simultaneous improvement across all OOD and ID metrics confirms that the three objectives provide complementary, non-conflicting supervision signals: $\mathcal{L}_{\mathrm{COV}}$ and $\mathcal{L}_{\mathrm{TEMP}}$ improve anomaly detection without sacrificing in-distribution classification.

As a continuation of Fig.~5 in the main paper, Fig.~\ref{fig:patient_case_study} presents additional representative longitudinal wound trajectories from four more patients spanning normal healing,  non-healing, and AE/SAE cases. For each clinical visit, we present the wound image together with the structured clinical context ($T_c$), including age, HbA1c, and UT grade, the corresponding wound description ($T_d$), and the fused OOD score produced by our multimodal framework. The examples illustrate how integrating visual appearance, clinical semantics, and temporal progression enables personalized wound assessment. Healing wounds exhibit progressive improvement in tissue appearance accompanied by consistently low OOD scores, whereas wounds with delayed healing or adverse events are expected to maintain elevated or increasing OOD scores, indicating greater deviation from the learned in-distribution healing patterns and an increased likelihood of clinically significant deterioration.



%% file: WACV_2027.bbl
\begin{thebibliography}{49}
\providecommand{\natexlab}[1]{#1}
\providecommand{\url}[1]{\texttt{#1}}
\expandafter\ifx\csname urlstyle\endcsname\relax
  \providecommand{\doi}[1]{doi: #1}\else
  \providecommand{\doi}{doi: \begingroup \urlstyle{rm}\Url}\fi

\bibitem[Anisuzzaman et~al.(2022{\natexlab{a}})Anisuzzaman, Wang, Rostami, Gopalakrishnan, Niezgoda, and Yu]{anisuzzaman2022multimodal}
D.~M. Anisuzzaman, Chuanbo Wang, Behrouz Rostami, Sandeep Gopalakrishnan, Jeffrey Niezgoda, and Zeyun Yu.
\newblock Multi-modal wound classification using wound image and location by deep neural network.
\newblock \emph{Scientific Reports}, 12\penalty0 (1):\penalty0 20057, 2022{\natexlab{a}}.

\bibitem[Anisuzzaman et~al.(2022{\natexlab{b}})Anisuzzaman, Wang, Rostami, Gopalakrishnan, Niezgoda, and Yu]{anisuzzaman2022review}
D.~M. Anisuzzaman, Chuanbo Wang, Behrouz Rostami, Sandeep Gopalakrishnan, Jeffrey Niezgoda, and Zeyun Yu.
\newblock Image-based artificial intelligence in wound assessment: A systematic review.
\newblock \emph{Advances in Wound Care}, 11\penalty0 (12):\penalty0 687--709, 2022{\natexlab{b}}.

\bibitem[Armstrong et~al.(2017)Armstrong, Boulton, and Bus]{armstrong2017diabetic}
David~G. Armstrong, Andrew J.~M. Boulton, and Sicco~A. Bus.
\newblock Diabetic foot ulcers and their recurrence.
\newblock \emph{New England Journal of Medicine}, 376\penalty0 (24):\penalty0 2367--2375, 2017.

\bibitem[Bannur et~al.(2023)Bannur, Hyland, Liu, Perez-Garcia, Ilse, Castro, Boecking, Sharma, Bouzid, Thieme, Schwaighofer, Wetscherek, Lungren, Nori, Alvarez-Valle, and Oktay]{bannur2023biovilt}
Shruthi Bannur, Stephanie Hyland, Qianchu Liu, Fernando Perez-Garcia, Maximilian Ilse, Daniel~C. Castro, Benedikt Boecking, Harshita Sharma, Kenza Bouzid, Anja Thieme, Anton Schwaighofer, Maria Wetscherek, Matthew~P. Lungren, Aditya Nori, Javier Alvarez-Valle, and Ozan Oktay.
\newblock Learning to exploit temporal structure for biomedical vision-language processing.
\newblock In \emph{Proceedings of the IEEE/CVF Conference on Computer Vision and Pattern Recognition (CVPR)}, pages 15016--15027, 2023.

\bibitem[Baur et~al.(2021)Baur, Denner, Wiestler, Navab, and Albarqouni]{baur2021autoencoders}
Christoph Baur, Stefan Denner, Benedikt Wiestler, Nassir Navab, and Shadi Albarqouni.
\newblock Autoencoders for unsupervised anomaly segmentation in brain {MR} images: A comparative study.
\newblock \emph{Medical Image Analysis}, 69:\penalty0 101952, 2021.

\bibitem[Boulton et~al.(2005)Boulton, Vileikyte, Ragnarson-Tennvall, and Apelqvist]{boulton2018}
Andrew J.~M. Boulton, Loretta Vileikyte, Gunnel Ragnarson-Tennvall, and Jan Apelqvist.
\newblock The global burden of diabetic foot disease.
\newblock \emph{The Lancet}, 366\penalty0 (9498):\penalty0 1719--1724, 2005.

\bibitem[Cassidy et~al.(2022)Cassidy, Kendrick, Reeves, Pappachan, O'Shea, Armstrong, and Yap]{cassidy2022dfuc2021}
Bill Cassidy, Connah Kendrick, Neil~D. Reeves, Joseph~M. Pappachan, Claire O'Shea, David~G. Armstrong, and Moi~Hoon Yap.
\newblock Diabetic foot ulcer grand challenge 2021: Evaluation and summary.
\newblock In \emph{Diabetic Foot Ulcers Grand Challenge (DFUC 2021), Lecture Notes in Computer Science, vol. 13183}. Springer, 2022.

\bibitem[Cruciani et~al.(2025)Cruciani, Kappelhof, Hartogsveld, Schoonmade, Roelofs, Peters, Smits, Wijnands, van Bavel, van~der Steen, and Schinkel]{cruciani2025dmwat}
Fabio Cruciani, Manon Kappelhof, Bram Hartogsveld, Linda Schoonmade, Joanne J. T.~H. Roelofs, Max J.~L. Peters, Ivo Smits, Roel Wijnands, Ed van Bavel, Anton F.~W. van~der Steen, and Michiel Schinkel.
\newblock Multimodal ai on wound images and clinical notes for home patient referral.
\newblock \emph{arXiv preprint arXiv:2501.13247}, 2025.

\bibitem[Djurisic et~al.(2023)Djurisic, Bozanic, Ashok, and Liu]{djurisic2023ash}
Andrija Djurisic, Nebojsa Bozanic, Arjun Ashok, and Rosanne Liu.
\newblock Extremely simple activation shaping for out-of-distribution detection.
\newblock In \emph{Proceedings of the International Conference on Learning Representations (ICLR)}, 2023.

\bibitem[Graham et~al.(2023)Graham, Pinaya, Tudosiu, Nachev, Ourselin, and Cardoso]{graham2023diffusion}
Mark~S. Graham, Walter H.~L. Pinaya, Petru-Daniel Tudosiu, Parashkev Nachev, Sebastien Ourselin, and Jorge Cardoso.
\newblock Denoising diffusion models for out-of-distribution detection.
\newblock In \emph{Proceedings of the IEEE/CVF Conference on Computer Vision and Pattern Recognition Workshops (CVPRW)}, pages 2948--2957, 2023.

\bibitem[Gu et~al.(2021)Gu, Tinn, Cheng, Lucas, Usuyama, Liu, Naumann, Gao, and Poon]{gu2021pubmedbert}
Yu Gu, Robert Tinn, Hao Cheng, Michael Lucas, Naoto Usuyama, Xiaodong Liu, Tristan Naumann, Jianfeng Gao, and Hoifung Poon.
\newblock Domain-specific language model pretraining for biomedical natural language processing.
\newblock \emph{ACM Transactions on Computing for Healthcare}, 3\penalty0 (1):\penalty0 1--23, 2021.

\bibitem[Hendrycks and Gimpel(2017)]{hendrycks2017msp}
Dan Hendrycks and Kevin Gimpel.
\newblock A baseline for detecting misclassified and out-of-distribution examples in neural networks.
\newblock In \emph{Proceedings of the International Conference on Learning Representations (ICLR)}, 2017.

\bibitem[Hendrycks et~al.(2019)Hendrycks, Mazeika, and Dietterich]{hendrycks2019deep}
Dan Hendrycks, Mantas Mazeika, and Thomas Dietterich.
\newblock Deep anomaly detection with outlier exposure.
\newblock In \emph{Proceedings of the International Conference on Learning Representations (ICLR)}, 2019.

\bibitem[Hong et~al.(2024)Hong, Yue, Chen, Cong, Lin, Luo, Wang, Wang, Xu, Yang, et~al.]{hong2024medood}
Ziyang Hong, Yizhou Yue, Yongkang Chen, Lin Cong, Hongbo Lin, Yuning Luo, Menghan Wang, Wei Wang, Jialin Xu, Xiaokun Yang, et~al.
\newblock Out-of-distribution detection in medical image analysis: A survey.
\newblock \emph{arXiv preprint arXiv:2404.18279}, 2024.

\bibitem[Houlsby et~al.(2019)Houlsby, Giurgiu, Jastrzebski, Morrone, de~Laroussilhe, Gesmundo, Attariyan, and Gelly]{houlsby2019adapter}
Neil Houlsby, Andrei Giurgiu, Stanislaw Jastrzebski, Bruna Morrone, Quentin de Laroussilhe, Andrea Gesmundo, Mona Attariyan, and Sylvain Gelly.
\newblock Parameter-efficient transfer learning for nlp.
\newblock In \emph{Proceedings of the International Conference on Machine Learning (ICML)}, pages 2790--2799, 2019.

\bibitem[Hu et~al.(2022)Hu, Shen, Wallis, Allen-Zhu, Li, Wang, Wang, and Chen]{hu2022lora}
Edward~J. Hu, Yelong Shen, Phillip Wallis, Zeyuan Allen-Zhu, Yuanzhi Li, Shean Wang, Lu Wang, and Weizhu Chen.
\newblock {LoRA}: Low-rank adaptation of large language models.
\newblock In \emph{International Conference on Learning Representations}, 2022.

\bibitem[Jeong et~al.(2023)Jeong, Zou, Kim, Zhang, Ravichandran, and Dabeer]{jeong2023winclip}
Jongheon Jeong, Yang Zou, Taewan Kim, Dongqing Zhang, Avinash Ravichandran, and Onkar Dabeer.
\newblock Winclip: Zero-/few-shot anomaly classification and segmentation.
\newblock In \emph{Proceedings of the IEEE/CVF Conference on Computer Vision and Pattern Recognition (CVPR)}, pages 19606--19616, 2023.

\bibitem[Jiang et~al.(2024)Jiang, Liu, Fang, Chen, Liu, Zheng, and Han]{jiang2024neglabel}
Xue Jiang, Feng Liu, Zhen Fang, Hong Chen, Tongliang Liu, Feng Zheng, and Bo Han.
\newblock Negative label guided ood detection with pretrained vision-language models.
\newblock In \emph{Proceedings of the International Conference on Learning Representations (ICLR)}, 2024.

\bibitem[Lee et~al.(2018)Lee, Lee, Lee, and Shin]{lee2018mahalanobis}
Kimin Lee, Kibok Lee, Honglak Lee, and Jinwoo Shin.
\newblock A simple unified framework for detecting out-of-distribution samples and adversarial attacks.
\newblock In \emph{Advances in Neural Information Processing Systems}, pages 7167--7177, 2018.

\bibitem[Li et~al.(2024)Li, Wong, Zhang, Usuyama, Liu, Yang, Naumann, Poon, and Gao]{li2024llava_med}
Chunyuan Li, Cliff Wong, Sheng Zhang, Naoto Usuyama, Haotian Liu, Jianwei Yang, Tristan Naumann, Hoifung Poon, and Jianfeng Gao.
\newblock Llava-med: Training a large language-and-vision assistant for biomedicine in one day.
\newblock In \emph{Advances in Neural Information Processing Systems (NeurIPS) Datasets and Benchmarks Track}, 2024.

\bibitem[Li et~al.(2023)Li, Li, Savarese, and Hoi]{li2023blip2}
Junnan Li, Dongxu Li, Silvio Savarese, and Steven Hoi.
\newblock {BLIP-2}: Bootstrapping language-image pre-training with frozen image encoders and large language models.
\newblock In \emph{International Conference on Machine Learning}, pages 19730--19742, 2023.

\bibitem[Li and Liang(2021)]{li2021prefix}
Xiang~Lisa Li and Percy Liang.
\newblock Prefix-tuning: Optimizing continuous prompts for generation.
\newblock In \emph{Proceedings of the 59th Annual Meeting of the Association for Computational Linguistics (ACL)}, pages 4582--4597, 2021.

\bibitem[Liang et~al.(2018)Liang, Li, and Srikant]{liang2018odin}
Shiyu Liang, Yixuan Li, and R. Srikant.
\newblock Enhancing the reliability of out-of-distribution image detection in neural networks.
\newblock In \emph{Proceedings of the International Conference on Learning Representations (ICLR)}, 2018.

\bibitem[Liu et~al.(2020)Liu, Wang, Owens, and Li]{liu2020energy}
Weitang Liu, Xiaoyun Wang, John Owens, and Yixuan Li.
\newblock Energy-based out-of-distribution detection.
\newblock In \emph{Advances in Neural Information Processing Systems (NeurIPS)}, pages 21464--21475, 2020.

\bibitem[Ming et~al.(2022{\natexlab{a}})Ming, Cai, Gu, Sun, Li, and Li]{ming2022delving}
Yifei Ming, Ziyang Cai, Jiuxiang Gu, Yiyou Sun, Wei Li, and Yixuan Li.
\newblock Delving into out-of-distribution detection with vision-language representations.
\newblock In \emph{Advances in Neural Information Processing Systems}, 2022{\natexlab{a}}.

\bibitem[Ming et~al.(2022{\natexlab{b}})Ming, Cai, Gu, Sun, Li, and Li]{ming2022ood}
Yifei Ming, Ziyang Cai, Jiuxiang Gu, Yiyou Sun, Wei Li, and Yixuan Li.
\newblock Delving into out-of-distribution detection with vision-language representations.
\newblock In \emph{Advances in Neural Information Processing Systems}, 2022{\natexlab{b}}.

\bibitem[Miyai et~al.(2023)Miyai, Yu, Irie, and Aizawa]{miyai2023locoop}
Atsuyuki Miyai, Qing Yu, Go Irie, and Kiyoharu Aizawa.
\newblock Locoop: Few-shot out-of-distribution detection via prompt learning.
\newblock In \emph{Advances in Neural Information Processing Systems (NeurIPS)}, pages 76298--76310, 2023.

\bibitem[Miyai et~al.(2025)Miyai, Yu, Irie, and Aizawa]{miyai2025glmcm}
Atsuyuki Miyai, Qing Yu, Go Irie, and Kiyoharu Aizawa.
\newblock Gl-mcm: Global and local maximum concept matching for zero-shot out-of-distribution detection.
\newblock \emph{International Journal of Computer Vision}, 2025.

\bibitem[Moor et~al.(2023)Moor, Hegselmann, Gruber, Kim, Klug, Bose, Katan, Nori, Davison, Bhatt, Fries, and Shah]{moor2023medflamingo}
Michael Moor, Stefan Hegselmann, Niklas Gruber, Hyun~Jin Kim, Michela Klug, Aditya Bose, Roy Katan, Harsha Nori, Matthew Davison, Urvashi Bhatt, Jason~Alan Fries, and Nigam~H. Shah.
\newblock Med-flamingo: A multimodal medical few-shot learner.
\newblock In \emph{Proceedings of Machine Learning for Health (ML4H)}, 2023.

\bibitem[Naiknaware and Sekeh(2026)]{naiknaware2026tqpm}
Aditi Naiknaware and Salimeh Sekeh.
\newblock {T-QPM}: Enabling temporal out-of-distribution detection and domain generalization for vision-language models in open-world.
\newblock \emph{arXiv preprint arXiv:2603.18481}, 2026.

\bibitem[Radford et~al.(2021)Radford, Kim, Hallacy, Ramesh, Goh, Agarwal, Sastry, Askell, Mishkin, Clark, Krueger, and Sutskever]{radford2021clip}
Alec Radford, Jong~Wook Kim, Chris Hallacy, Aditya Ramesh, Gabriel Goh, Sandhini Agarwal, Girish Sastry, Amanda Askell, Pamela Mishkin, Jack Clark, Gretchen Krueger, and Ilya Sutskever.
\newblock Learning transferable visual models from natural language supervision.
\newblock In \emph{Proceedings of the 38th International Conference on Machine Learning}, pages 8748--8763, 2021.

\bibitem[Rostami et~al.(2021)Rostami, Anisuzzaman, Wang, Gopalakrishnan, Niezgoda, and Yu]{rostami2021ensemble}
Behrouz Rostami, D.~M. Anisuzzaman, Chuanbo Wang, Sandeep Gopalakrishnan, Jeffrey Niezgoda, and Zeyun Yu.
\newblock Multiclass wound image classification using an ensemble deep cnn-based classifier.
\newblock \emph{Computers in Biology and Medicine}, 134:\penalty0 104536, 2021.

\bibitem[Scebba et~al.(2022)Scebba, Zhang, Catanzaro, Mihai, Distler, Berli, and Karlen]{scebba2022wound}
Gaetano Scebba, Jia Zhang, Sara Catanzaro, Cosmin Mihai, Oliver Distler, Martin Berli, and Walter Karlen.
\newblock Detect-and-segment: A deep learning approach to automate wound image segmentation.
\newblock \emph{Informatics in Medicine Unlocked}, 29:\penalty0 100884, 2022.

\bibitem[Sekeh and Wisell(2026)]{sekeh2026crossmodal}
Salimeh Sekeh and Mary Wisell.
\newblock Understanding cross-modal contributions in continual vision-language models: A theoretical perspective.
\newblock \emph{arXiv preprint arXiv:2606.14883}, 2026.

\bibitem[Sun et~al.(2021)Sun, Guo, and Li]{sun2021react}
Yiyou Sun, Chuan Guo, and Yixuan Li.
\newblock React: Out-of-distribution detection with rectified activations.
\newblock In \emph{Advances in Neural Information Processing Systems (NeurIPS)}, pages 144--157, 2021.

\bibitem[Sun et~al.(2022)Sun, Ming, Zhu, and Li]{sun2022knnood}
Yiyou Sun, Yifei Ming, Xiaojin Zhu, and Yixuan Li.
\newblock Out-of-distribution detection with deep nearest neighbors.
\newblock In \emph{Proceedings of the International Conference on Machine Learning (ICML)}, pages 20827--20840, 2022.

\bibitem[Tack et~al.(2020)Tack, Mo, Jeong, and Shin]{tack2021csi}
Jihoon Tack, Sangwoo Mo, Jongheon Jeong, and Jinwoo Shin.
\newblock Csi: Novelty detection via contrastive learning on distributionally shifted instances.
\newblock In \emph{Advances in Neural Information Processing Systems (NeurIPS)}, pages 11839--11852, 2020.

\bibitem[Vaswani et~al.(2017)Vaswani, Shazeer, Parmar, Uszkoreit, Jones, Gomez, Kaiser, and Polosukhin]{vaswani2017attention}
Ashish Vaswani, Noam Shazeer, Niki Parmar, Jakob Uszkoreit, Llion Jones, Aidan~N. Gomez, Lukasz Kaiser, and Illia Polosukhin.
\newblock Attention is all you need.
\newblock In \emph{Advances in Neural Information Processing Systems (NeurIPS)}, 2017.

\bibitem[Wang et~al.(2020)Wang, Anisuzzaman, Williamson, Dhar, Rostami, Niezgoda, Gopalakrishnan, and Yu]{wang2020woundseg}
Chuanbo Wang, D.~M. Anisuzzaman, Victor Williamson, Mrinal~Kanti Dhar, Behrouz Rostami, Jeffrey Niezgoda, Sandeep Gopalakrishnan, and Zeyun Yu.
\newblock Fully automatic wound segmentation with deep convolutional neural networks.
\newblock \emph{Scientific Reports}, 10\penalty0 (1):\penalty0 21897, 2020.

\bibitem[Wang et~al.(2022{\natexlab{a}})Wang, Li, Feng, and Zhang]{zhang2022vim}
Haoqi Wang, Zhizhong Li, Litong Feng, and Wayne Zhang.
\newblock Vim: Out-of-distribution with virtual-logit matching.
\newblock In \emph{Proceedings of the IEEE/CVF Conference on Computer Vision and Pattern Recognition (CVPR)}, pages 4921--4930, 2022{\natexlab{a}}.

\bibitem[Wang et~al.(2023)Wang, Li, Yao, and Li]{wang2023clipn}
Hualiang Wang, Yi Li, Huifeng Yao, and Xiaomeng Li.
\newblock Clipn for zero-shot ood detection: Teaching clip to say no.
\newblock In \emph{Proceedings of the IEEE/CVF International Conference on Computer Vision (ICCV)}, pages 1802--1812, 2023.

\bibitem[Wang et~al.(2022{\natexlab{b}})Wang, Wu, Agarwal, and Sun]{wang2022medclip}
Zifeng Wang, Zhenbang Wu, Dinesh Agarwal, and Jimeng Sun.
\newblock Medclip: Contrastive learning from unpaired medical images and text.
\newblock \emph{arXiv preprint arXiv:2210.10163}, 2022{\natexlab{b}}.

\bibitem[Yang et~al.(2022)Yang, Wang, Zou, Zhou, Ding, Peng, Wang, Chen, Li, Sun, Du, Zhou, Zhang, Hendrycks, Li, and Liu]{yang2022openood}
Jingkang Yang, Pengyun Wang, Dejian Zou, Zitang Zhou, Kunyuan Ding, Wenxuan Peng, Haoqi Wang, Guangyao Chen, Bo Li, Yiyou Sun, Xuefeng Du, Kaiyang Zhou, Wayne Zhang, Dan Hendrycks, Yixuan Li, and Ziwei Liu.
\newblock Openood: Benchmarking generalized out-of-distribution detection.
\newblock In \emph{Advances in Neural Information Processing Systems (NeurIPS) Datasets and Benchmarks Track}, pages 32598--32611, 2022.

\bibitem[Yang et~al.(2024)Yang, Zhou, Li, and Liu]{yang2024generalized}
Jingkang Yang, Kaiyang Zhou, Yixuan Li, and Ziwei Liu.
\newblock Generalized out-of-distribution detection: A survey.
\newblock \emph{International Journal of Computer Vision}, 132\penalty0 (12):\penalty0 5635--5662, 2024.

\bibitem[Zhang et~al.(2023{\natexlab{a}})Zhang, Chen, Bukharin, Karampatziakis, He, Cheng, Chen, and Zhao]{hu2023adalora}
Qingru Zhang, Minshuo Chen, Alexander Bukharin, Nikos Karampatziakis, Pengcheng He, Yu Cheng, Weizhu Chen, and Tuo Zhao.
\newblock Adalora: Adaptive budget allocation for parameter-efficient fine-tuning.
\newblock In \emph{Proceedings of the International Conference on Learning Representations (ICLR)}, 2023{\natexlab{a}}.

\bibitem[Zhang et~al.(2023{\natexlab{b}})Zhang, Xu, Usuyama, Bagga, Tinn, Preston, Rao, Wei, Valluri, Wong, Lungren, Naumann, and Poon]{zhang2023biomedclip}
Sheng Zhang, Yanbo Xu, Naoto Usuyama, Jaspreet~Kaur Bagga, Robert Tinn, Sam Preston, Rajesh Rao, Mu-Hsin Wei, Naveen Valluri, Cliff Wong, Matthew~P. Lungren, Tristan Naumann, and Hoifung Poon.
\newblock {BiomedCLIP}: a multimodal biomedical foundation model pretrained from fifteen million scientific image-text pairs.
\newblock \emph{arXiv preprint arXiv:2303.00915}, 2023{\natexlab{b}}.

\bibitem[Zhang et~al.(2024)Zhang, Xu, and Xiang]{zhang2024vision}
Zihan Zhang, Zhuo Xu, and Xiang Xiang.
\newblock Vision-language dual-pattern matching for out-of-distribution detection.
\newblock In \emph{European Conference on Computer Vision}, pages 273--291. Springer, 2024.

\bibitem[Zhou et~al.(2022{\natexlab{a}})Zhou, Yang, Loy, and Liu]{zhou2022cocoop}
Kaiyang Zhou, Jingkang Yang, Chen~Change Loy, and Ziwei Liu.
\newblock Conditional prompt learning for vision-language models.
\newblock In \emph{IEEE/CVF Conference on Computer Vision and Pattern Recognition}, pages 16816--16825, 2022{\natexlab{a}}.

\bibitem[Zhou et~al.(2022{\natexlab{b}})Zhou, Yang, Loy, and Liu]{zhou2022coop}
Kaiyang Zhou, Jingkang Yang, Chen~Change Loy, and Ziwei Liu.
\newblock Learning to prompt for vision-language models.
\newblock \emph{International Journal of Computer Vision}, 130\penalty0 (9):\penalty0 2337--2348, 2022{\natexlab{b}}.

\end{thebibliography}
